\documentclass[sigconf,natbib=true]{acmart}
\AtBeginDocument{%
  }

\setcopyright{acmlicensed}
\copyrightyear{2025}
\acmYear{2025}
\acmDOI{XXXXXXX.XXXXXXX}
\acmConference[Conference acronym 'XX]{Make sure to enter the correct
  conference title from your rights confirmation email}{June 03--05,
  2018}{Woodstock, NY}
\acmISBN{978-1-4503-XXXX-X/18/06}

\usepackage{arydshln}
\usepackage{lipsum}
\usepackage{tabularray}
\usepackage{multirow}
\usepackage{booktabs, caption, array}
\usepackage{algorithm}
\usepackage{algorithmic}
\usepackage{comment}
\usepackage{xcolor}
\usepackage{subcaption}
\usepackage{graphicx}
\usepackage[skins, breakable]{tcolorbox}
\usepackage{subcaption}
\usepackage{enumitem}
\usepackage{framed}
\usepackage{hyperref} 
\usepackage{amsmath}
\definecolor{shadecolor}{gray}{0.9}
\usepackage{comment}
\usepackage{xspace}
\usepackage{colortbl}
\usepackage{bbding}
\newcommand{\mymodel}{ReARTeR\xspace}
\definecolor{customgreen}{HTML}{b7d5a3}
\definecolor{customred}{HTML}{f17e82}
\definecolor{customblue}{HTML}{b4caf3}

\newtcolorbox{cvbox}[1][]{
  enhanced,
  breakable,
  colback=gray!20,
  colframe=gray,
  coltitle=white,
  fonttitle=\bfseries,
  boxsep=0mm,
  #1
}

\begin{document}

\title{\mymodel: Retrieval-Augmented Reasoning with Trustworthy Process Rewarding}

\author{Zhongxiang Sun}
\author{Qipeng Wang}
\affiliation{
  \institution{\mbox{Gaoling School of Artificial
Intelligence}\\Renmin University of China}
  \city{Beijing}\country{China}
  }
\email{{sunzhongxiang, wqp}@ruc.edu.cn}

\author{Weijie Yu}
\affiliation{
  \institution{\mbox{School of Information Technology} \\ and Management\\University of International Business and Economics
}
  \city{Beijing}\country{China}
  }
\email{yu@uibe.edu.cn}

\author{Xiaoxue Zang}
\author{Kai Zheng}
\affiliation{%
  \institution{Kuaishou Technology Co., Ltd.}
  \city{Beijing}\country{China}
  }
\email{xxic666@126.com}
\email{zhengkai@kuaishou.com}

\author{Jun Xu}
\authornote{Corresponding author. \\ Work partially done at Engineering Research Center of Next-Generation Intelligent Search and Recommendation, Ministry of Education.}
\author{Xiao Zhang}

\affiliation{%
  \institution{\mbox{Gaoling School of Artificial
Intelligence}\\Renmin University of China}
  \city{Beijing}\country{China}
  }
\email{{junxu,zhangx89}@ruc.edu.cn}

\author{Yang Song}
\author{Han Li}
\affiliation{%
  \institution{Kuaishou Technology Co., Ltd.}
  \city{Beijing}\country{China}
  }
\email{lihan08@kuaishou.com}
\email{ys@sonyis.me}

\renewcommand{\shortauthors}{Trovato et al.}

\begin{abstract}
Retrieval-Augmented Generation (RAG) systems for Large Language Models (LLMs) have shown promise in knowledge-intensive tasks, yet their reasoning capabilities, particularly for complex multi-step reasoning, remain limited. Although recent approaches have explored integrating RAG with chain-of-thought reasoning or incorporating test-time search with process reward model (PRM), these methods face several untrustworthy challenges, including lack of explanations, bias in PRM training data, early-step bias in PRM scores, and ignoring post-training that fails to fully optimize reasoning potential.
To address these issues, we propose \textbf{Re}trieval-\textbf{A}ugmented \textbf{R}easoning through \textbf{T}rustworthy Proc\textbf{e}ss \textbf{R}ewarding (\textbf{\mymodel}), a framework that enhances RAG systems' reasoning capabilities through both post-training and test-time scaling. At test time, \mymodel introduces Trustworthy Process Rewarding via a Process Reward Model for accurate scalar scoring and a Process Explanation Model (PEM) for generating natural language explanations, enabling step refinement. During post-training, we leverage Monte Carlo Tree Search guided by Trustworthy Process Rewarding to collect high-quality step-level preference data, which is used to optimize the model through Iterative Preference Optimization.
\mymodel tackles three key challenges: (1) misalignment between PRM and PEM, addressed through off-policy preference learning; (2) bias in PRM training data, mitigated by a balanced annotation method and incorporating stronger annotations for difficult examples; and (3) early-step bias in PRM, resolved via a temporal-difference-based look-ahead search strategy.
Experimental results on multi-step reasoning benchmarks demonstrate that \mymodel significantly improves reasoning performance, highlighting its potential to advance the reasoning capability of RAG systems.
\end{abstract}




\maketitle

\section{Introduction}
Retrieval-augmented generation (RAG) for Large Language Models (LLMs) is widely utilized to address knowledge-intensive tasks, typically comprising a generator (LLM) and a retriever (for external knowledge retrieval)~\cite{xu2024recomp,adaptiverag,kim2024sure,ragsurvey}. However, complex multi-step reasoning tasks remain challenging even for the most advanced RAG systems. Existing works have explored integrating RAG with chain-of-thought (CoT) reasoning~\cite{trivedi2023interleaving,yao2023react}, but these approaches have yet to fully leverage the reasoning potential of LLMs. 

\begin{figure}[t]
    \centering
    \includegraphics[width=\linewidth]{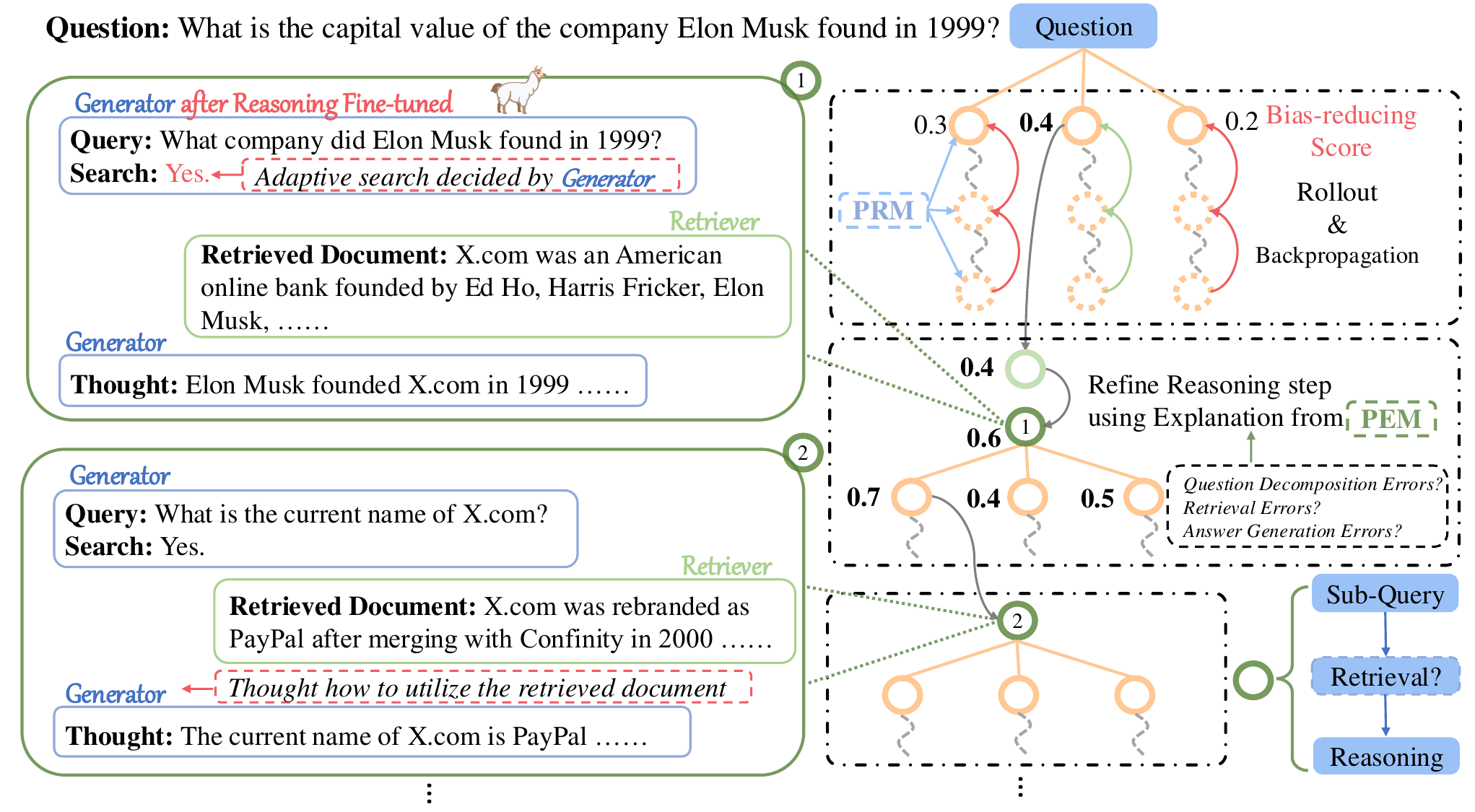}
    \caption{An example of how RARTPR tackles complex multi-step questions. The right part highlights test-time search with PRM and refinement via PEM explanations, while the left part details the reasoning step, including sub-query, adaptive retrieval, and reasoning thought.}
    \label{fig:intro}
\end{figure}

Recently, Process Reward Models (PRMs) have been introduced to enhance the reasoning capability of RAG systems through test-time scaling~\cite{li2024can,dong2024progressive}. However, these methods often face untrustworthy challenges:
\textbf{(1) Lack of Explanations:} Existing PRMs often generate unexplainable scalar scores and cannot incorporate natural language critiques, which limits interpretability and hinders their effectiveness in enhancing refinement during test-time reasoning~\cite{luo2024improve,wang2024math};  
\textbf{(2) Bias in PRM training data:} Traditional Monte Carlo methods for collecting Process Supervision Datasets often result in a distributional bias, where most questions receive disproportionately high scores~\cite{li2024can, wang2024math, wang2024multi}. Consequently, the PRM struggles to identify erroneous steps and fails to provide meaningful feedback on difficult examples; 
\textbf{(3) Early-Step Bias in PRM:} PRMs exhibit reduced accuracy in predicting rewards for earlier reasoning steps compared to those closer to the reasoning endpoint, due to the increased randomness and uncertainty in earlier steps; 
\textbf{(4) Lack of Reasoning Optimization:} Additionally, these approaches rely on off-the-shelf LLMs as generators without incorporating reasoning-specific optimization during the post-training phase~\cite{luong2024reft,zhang2024openrft,xie2024monte}.

To address the above limitations and improve the reasoning capabilities of RAG systems, we explore enhancing \textbf{R}etrieval-\textbf{A}ugmented \textbf{R}easoning through \textbf{T}rustworthy  \textbf{P}rocess \textbf{R}ewarding (\textbf{\mymodel}) in both test-time and post-training scaling. Specifically, as shown in~\autoref{fig:intro}, the \textbf{testing phase} is guided by the Trustworthy Process Rewarding. The Trustworthy Process Rewarding is implemented through two models: (1) a Process Reward Model (\textbf{PRM}), which provides scalar scores that, while accurate, lack interpretability; and (2) a Process Explanation Model (\textbf{PEM}), which generates natural language explanations for the process reward model's scores, facilitating refinement of steps with lower scores. 
During the \textbf{post-training phase}, we introduce step-level offline reinforcement fine-tuning to enhance the reasoning capabilities of the RAG system. Specifically,  recognizing the dynamic interaction between the generator and retriever in RAG, on each iteration we employ Monte Carlo Tree Search (MCTS)~\cite{browne2012survey} guided by Trustworthy Process Rewarding to generate high-quality, step-level preference data. This data is subsequently utilized to optimize the model, resulting in a substantial improvement in the system's reasoning performance.

As the core component of \mymodel, the Trustworthy Process Rewarding solving the following challenges:
\textbf{(1) Misalignment between the PEM and PRM:} Off-the-shelf LLMs used as PEM often generate explanations that are not aligned with the PRM's scalar scores, hindering the generator's ability to refine outputs based on external feedback. To address this issue, we propose aligning the PEM with the PRM through Off-policy Preference Learning, which leverages preference labels derived from PRM scores before and after the generator refines the reasoning step based on PEM explanations. If the explanation improves the PRM score, it is treated as a positive example; otherwise, it is treated as a negative example;
\textbf{(2) Bias in PRM training data:}  To mitigate this, we leverage OmegaPRM~\cite{luo2024improve}, which emphasizes identifying errors in reasoning steps and balances positive and negative examples. For challenging samples, we incorporate annotations from stronger models or human experts to provide accurate reasoning steps, thereby enhancing the PRM’s ability to discern correct reasoning paths in difficult scenarios;
\textbf{(3) Early-Step Bias in PRM:}  To resolve this, we propose a temporal-difference (TD)-based look-ahead search strategy, where simulated future reasoning steps are used to compute expected rewards, enabling updates to the current step's reward estimation. Compared to previous approaches~\cite{snell2024scaling}, this method effectively achieves a balance between bias and variance.

We summarize the major contributions of this paper as follows:

(1) We pioneer the exploration of combining post-training and test-time scaling to enhance reasoning capabilities in Retrieval-Augmented Generation scenarios. By integrating Trustworthy Process Rewarding, \mymodel significantly improves the quality of reasoning paths discovered during the post-training phase, as well as the accuracy of search and the generator's refinement ability during the test phase.

(2) We tackle key challenges in implementing Trustworthy Process Rewarding by aligning the PEM and PRM through off-policy preference learning, balancing the training data of PRM, and employing a TD-based look-ahead search strategy to reduce Early-Step Bias of PRM.

(3) Experimental results demonstrate that \mymodel achieves significant improvements on multiple public multi-step reasoning RAG datasets, validating the feasibility of enhancing RAG systems' reasoning capabilities through post-training and inference-time scaling with \mymodel.





\section{Related Work}

\subsection{Learning and Search for Reasoning}
Advanced reasoning models often follow the \textit{learning and search} principle~\cite{sutton2019bitter} to enhance reasoning capabilities through post-training and test-time scaling strategies.

\textit{Post-training Scaling.} 
ReFT~\cite{luong2024reft} employs reinforcement fine-tuning, where LLMs explore reasoning paths and optimize based on feedback, using PPO for training. While PPO achieves better results than DPO due to interactive updates, it suffers from instability. Iterative training methods~\cite{singh2023beyond} offer more stability and efficiency, with Iterative Preference Optimization~\cite{pang2024iterative} improving reasoning by constructing preference CoT data and using iterative DPO. However, these approaches face challenges in collecting step-level reasoning preferences and rely on difficult-to-collect pairwise data. To address these limitations, we propose using MCTS to collect step-level preference data and employ KTO~\cite{ethayarajh2024kto} for stable optimization, leveraging process supervision to enhance reasoning.

\textit{Test-time Scaling.} 
Test-time scaling typically relies on (1) \textbf{Self-Refinement}, where models iteratively improve outputs~\cite{madaan2024self}, but this approach is limited by the lack of external feedback~\cite{huang2024large}; and (2) \textbf{Search with Verifier}, which generates multiple outputs and selects the best using a verifier, such as a Process Reward Model (PRM). While PRM scores have been used as feedback for Self-Refinement~\cite{wu2024enhancing}, they often fail to guide effective improvements in RAG scenarios. To overcome this, we combine PRM-aligned PEM explanations with step-level Self-Refinement for better reasoning performance.
PRMs are critical during search, but their training data collection significantly affects performance. Existing methods~\cite{li2024can,dong2024progressive} use Monte Carlo methods to generate process supervision signals, discarding reasoning steps after rollouts, resulting in inefficiency. OmegaPRM~\cite{luo2024improve} improves this by storing rollouts for reuse and using binary search to identify errors, balancing positive and negative examples. Building on OmegaPRM, we incorporate stronger generators for difficult problems and propose a TD-based look-ahead search strategy to enhance PRM accuracy for shallow reasoning nodes, achieving trustworthy process rewards.

\subsection{Retrieval-Augmented Reasoning}

Retrieval-augmented generation for Large Language Models is widely used for knowledge-intensive tasks~\cite{xu2024recomp,adaptiverag,kim2024sure,ragsurvey}, but remains limited in handling complex multi-step reasoning. Facing this challenge, existing works integrate RAG with CoT reasoning~\cite{trivedi2023interleaving,yao2023react}. For instance, Self-Ask~\cite{press2022measuring} uses CoT to explicitly reason through follow-up questions before addressing the query, while IRCoT~\cite{trivedi2023interleaving} interleaves retrieval with reasoning steps to iteratively refine reasoning using CoT and retrieved results. Recently, \citet{yue2024inference} proposed an iterative demonstration-based RAG method that performs multiple iterations to achieve test-time scaling. However, these approaches primarily leverage the long context capabilities of LLMs and directly combine CoT with retrieval without effectively utilizing learning and search to enhance the reasoning capabilities of RAG systems.
CR-Planner~\cite{li2024can} attempts to directly use Process Reward Models (PRMs) to assist search and improve the reasoning capability of RAG systems through test-time scaling. However, it fails to address the untrustworthy challenges inherent in PRMs.
We pioneer the use of trustworthy process rewarding to guide both post-training scaling and test-time scaling, significantly enhancing the multi-step reasoning capabilities of RAG systems.

\begin{figure*}[t]
    \centering
    \includegraphics[width=\linewidth]{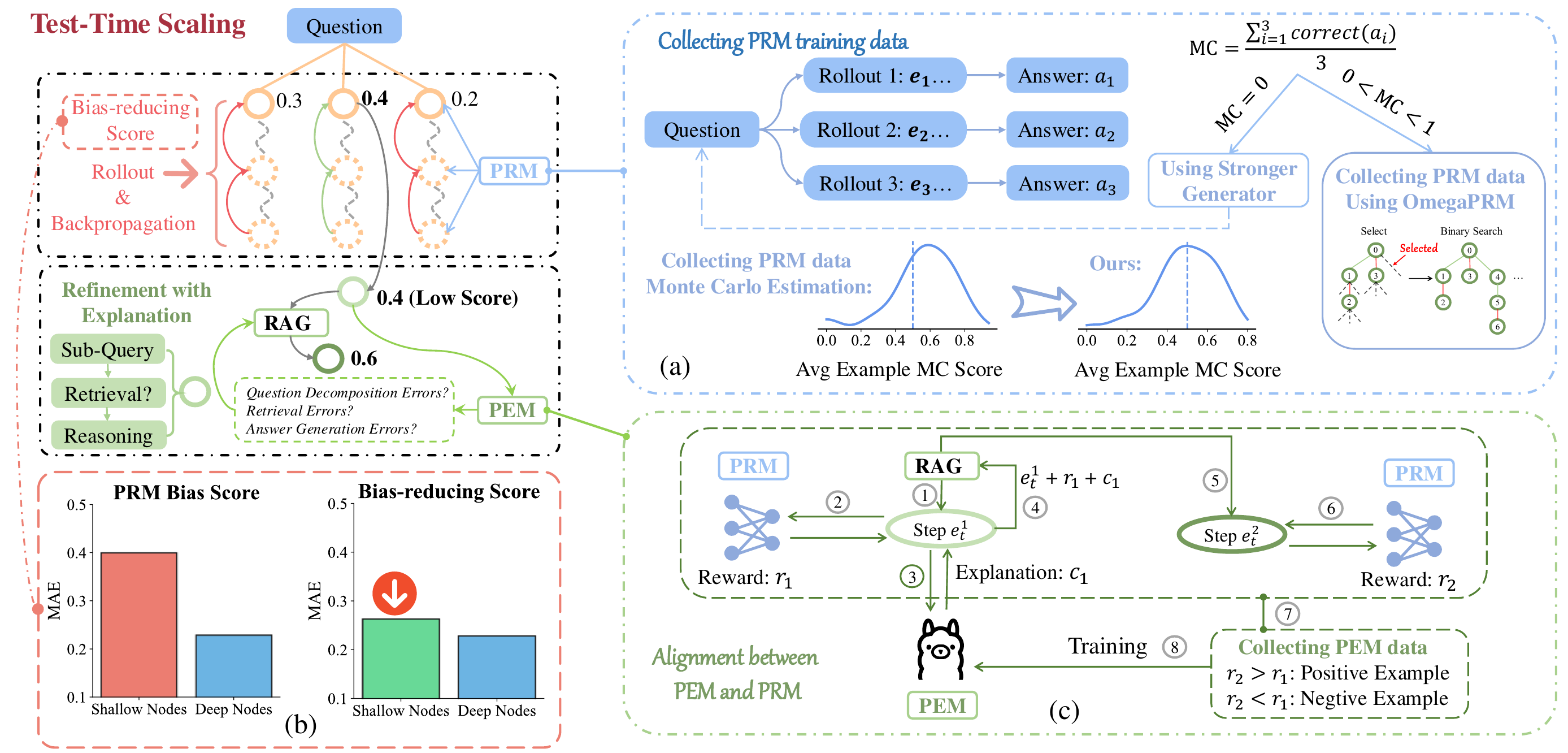}
    \caption{Test-Time Scaling of \mymodel, which includes \textcolor{customblue}{collecting unbiased PRM training data for PRM}, \textcolor{customred}{reducing early-step bias in PRM}, and \textcolor{customgreen}{alignment between PEM and PRM}.}
    \label{fig:method_1}
\end{figure*}

\begin{figure*}[t]
    \centering
    \includegraphics[width=\linewidth]{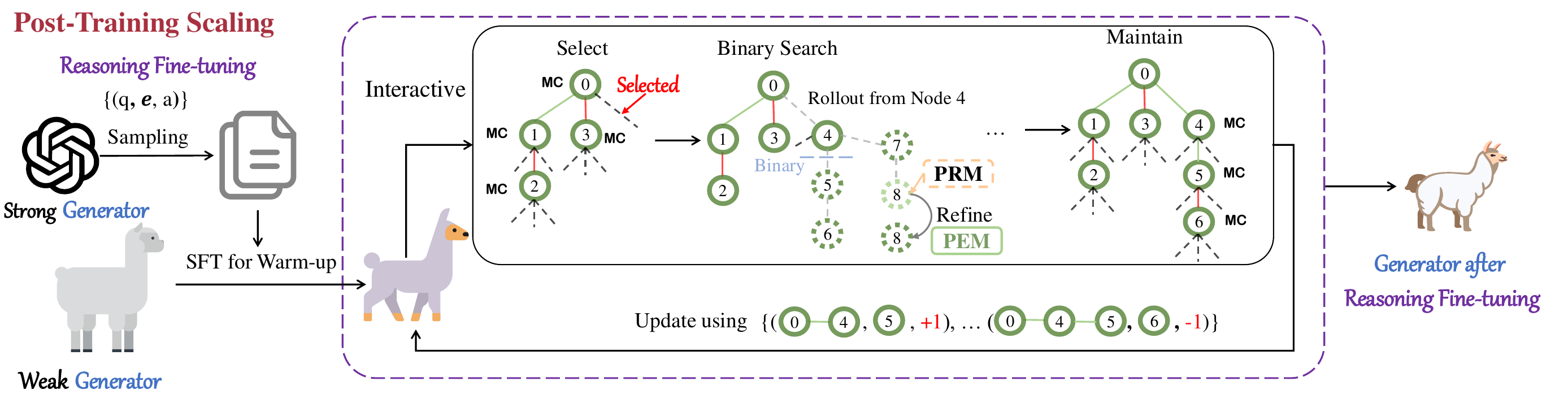}
    \caption{Post-Training Scaling of \mymodel, which includes Warm-Up and Step-Level Offline Reinforcement Stages.}
    \label{fig:method_2}
\end{figure*}

\section{Method}
\subsection{Overview}

In this section, we present the overview of \mymodel, which enhances Retrieval-Augmented Reasoning through Trustworthy Process Rewarding in both test-time and post-training scaling. The policy model $\pi_{\theta}$ of \mymodel includes a generator $G$, which can either be an off-the-shelf LLM such as the proprietary model GPT4-o~\cite{achiam2023gpt} or an open-source model such as LLaMA3~\cite{dubey2024llama} which can be post-trained for enhancing reasoning, and a retriever $E$. Additionally, \mymodel incorporates a Process Reward Model (PRM) $R$ and a Process Explanation Model (PEM) $C$.

Given a complex multi-step question $q$ and a retrieval corpus $\mathcal{D}$, \mymodel generates a reasoning process (CoT) $\mathbf{e}$ before producing an answer $a$ to $q$. The CoT of \mymodel consists of a sequence of reasoning steps:
\begin{equation}
\mathbf{e} = [e_1, e_2, \ldots, e_T],
\end{equation}
where $T$ represents the maximum length of the reasoning steps. 

As illustrated in~\autoref{fig:method_1}, each reasoning step $e_t$ comprises a sub-query $q_t$, a retrieval indicator $j_t$, external knowledge $d_t$ retrieved by $E$ from the corpus $\mathcal{D}$ if $j_t = \text{``Yes''}$, and a thought $r_t$ generated by the generator based on the context:
\begin{equation}
e_t = [q_t, j_t, d_t, r_t].
\label{eq:et}
\end{equation}

At timestep $t$, the reasoning step $e_t$ is sampled from the policy $\pi_\theta(s_t)$, where the state $s_t$ represents the combination of the question $q$ and the sequence of reasoning steps up to $e_{t-1}$.

For the sampling process of $e_t$, we first sample $M$ different reasoning steps:
\[
\mathcal{E}_t = [e_t^{1}, e_t^{2}, \ldots, e_t^{M}].
\]
Subsequently, the PRM $R$ predicts scores to the reasoning steps in $\mathcal{E}_t$:
\[
r_t = R(s_t, e_t), \quad r_t \in (0, 1).
\]
The reasoning step with the highest reward score is selected:
\[
\hat{e}_t = \arg\max_{e_t^m \in \mathcal{E}_t} R(s_t, e_t^m),
\]
if $\hat{r}_t > \tau$, then $e_t \leftarrow \hat{e}_t$ is directly added to $s_t$. Otherwise, a \textbf{refinement phase} is initiated, where the process critic model $C$ provides an explanation $c_t$ for the low process reward score of $\hat{e}_t$:
\[
c_t = C(s_t, \hat{e}_t, \hat{r}_t).
\]
The policy model then utilizes external feedback to correct $\hat{e}_t$:
\[
e_t = \pi_\theta(s_t | \hat{e}_t, c_t, \hat{r}_t).
\]
Finally, $e_t$ is added to the reasoning process:
\[
s_{t+1} = [s_t, e_t].
\]

Ultimately, the policy $\pi_\theta$ generates the final answer $a$ based on the question $q$ and the complete reasoning process $\mathbf{e}$.

In the following sections, we will introduce the implementation of the PRM~(\textbf{\textsection~\ref{sec:prm_train}}), including its training process and the method for reducing early-step bias for PRM. Additionally, we describe the training process of the PEM~(\textbf{\textsection~\ref{sec:pem_train}}) and the post-training scaling strategy for \mymodel~(\textbf{\textsection~\ref{sec:post_train}}).

\subsection{Process Reward Model Training}
\label{sec:prm_train}

The Process Reward Model of \mymodel is trained to truthfully predict the process reward score of each intermediate step $e_t$. 

\textbf{Training data collection:} Considering the training data requires process supervision labels which are hard to annotate, to reduce human annotation costs, existing methods propose an automatic annotation approach using the Monte Carlo method to generate process supervision signals~\cite{li2024can,dong2024progressive}. For each step of a CoT $\mathbf{e}$, multiple complete reasoning paths and final answers are obtained via rollouts. By evaluating the accuracy of the final answers, process supervision signals for the current reasoning step can be derived.

However, as shown in~\autoref{fig:method_1}(a), we observed that this method often introduces distributional bias, where most questions receive disproportionately high scores. Additionally, for difficult questions, the sampled process supervision signals frequently result in a value of zero, leaving the PRM unable to identify erroneous steps or provide meaningful feedback on challenging examples.

To address this issue, as illustrated in~\autoref{fig:method_1}(a), we first perform $N$ rollouts for the question $q$ to obtain $\{(q_1, \mathbf{e}_1, a_{1}), \ldots, (q_N, \mathbf{e}_N, a_{N})\}$. The accuracy of the final answers across all rollouts is used to compute the Monte Carlo (MC) score:
\begin{equation}
\text{MC} = \frac{\sum_{n=1}^{N} \text{correct}(a_{n})}{N}.
\label{eq:mc}
\end{equation}

For questions where $0 < \text{MC} < 1$, we employ the OmegaPRM~\cite{luo2024improve} annotation scheme, which efficiently identifies the first error in $\mathbf{e}$ using binary search and balances positive and negative examples, thereby ensuring both efficiency and quality. For questions where $\text{MC} = 0$, we switch to a stronger generator for reasoning. Questions with final $\text{MC} = 1$ or $\text{MC} = 0$ (even when using a stronger generator) are discarded, as they lack discriminative value and do not enable the model to identify correct or incorrect reasoning steps for specific questions.

For the selected questions, following the above process, we construct the process supervision data $\mathcal{D}_{\text{prm}} = \{(s_{i}, e_{i}, \text{MC}_{i})\}_{i=1}^{M_{r}}$, where $M_{r}$ represents the number of samples in $\mathcal{D}_{\text{prm}}$, and $\text{MC}_{i}$ is the MC score computed for $[s_{i}, e_{i}]$ after $N$ rollouts using Eq.~\ref{eq:mc}.

\textbf{PRM Training:}  
For each process supervision data $(s_{i}, e_{i}, \text{MC}_{i})$ in $\mathcal{D}_{\text{prm}}$, we define binary labels $y_i = 1$ if $\text{MC}_i > 0.5$, otherwise $y_i = 0$. We utilize the Cross-Entropy (CE) loss to train the PRM:
\[
\mathcal{L}_{\text{prm}} = -\frac{1}{M_r} \sum_{i=1}^{M_{r}} \left[ y_i \log R(s_i, e_i) + (1 - y_i) \log (1 - R(s_i, e_i)) \right],
\]
where $R$ denotes the process reward model.

\subsection{Reducing Early-Step Bias for PRM} At the inference stage of PRM, we observe that PRMs exhibit reduced accuracy in predicting rewards for earlier reasoning steps (shallow nodes) compared to those closer to the reasoning end-point (deep nodes), as shown in~\autoref{fig:method_1}(b). This phenomenon, attributed to the increased randomness and uncertainty in earlier steps, is referred to as \textbf{early-step bias}. 

Some existing works adopt a Lookahead Search strategy~\cite{snell2024scaling}, which performs a simulation by rolling out up to $H$ steps further, stopping early if the solution end-point is reached. The PRM’s score at the end of this rollout is then used to evaluate the current step during beam search. While this approach mitigates bias, it introduces significant variance~\cite{sutton2018reinforcement}. To achieve a bias-variance trade-off, inspired by Temporal Difference (TD) learning~\cite{td}, we propose a TD-based Lookahead Search to update the PRM scores for shallow nodes:
\[
r_t \leftarrow r_t + \alpha \Delta_t,
\]
where 
\[
\Delta_t = \left( r_{t+1} - r_t \right),
\]
$r_t = R(s_t, e_t)$, and $\alpha$ is the discount factor.

In our approach, the termination of the Lookahead Search simulation is adaptively determined by whether $\Delta_t$ falls below a threshold $\beta$ (indicating diminishing returns in further rollouts when reward scores stabilize) or if the predefined step limit $H$ is reached. This adaptive simulation mechanism balances computational efficiency and bias reduction, saving resources while maintaining performance.

\subsection{Process Explanation Model}
\label{sec:pem_train}

In this section, we introduce the training procedure of the Process Explanation Model. After training the PRM, the PRM can effectively score the reasoning process; however, the PRM score is an unexplainable scalar and cannot provide natural language critiques. To address this limitation, we designed PEM, a generative model specifically aimed at producing explanations for refinement. 

However, directly using off-the-shelf LLMs as PEM often results in explanations misaligned with the PRM’s scalar scores, hindering the generator’s ability to refine reasoning steps based on external feedback. 
To address this issue, as shown in~\autoref{fig:method_1}(c), we propose aligning the PEM with the PRM through \textbf{Off-policy Preference Learning}. This method uses the PRM as a verifier to provide feedback for the PEM-generated explanations, yielding preference data $\mathcal{D}_{\text{pem}}$. The PEM is then updated to align its explanations with the PRM’s scoring, facilitating the generation of explanations that enhance the policy model’s reasoning step through refinement.

Given the state $s_t$ and reasoning step $e_{t}^{1}$, the process is as follows:

1. The PRM provides an initial score:
   \[
   r_{t}^1 = R(s_t, e_t^{1}),
   \]
   and the PEM generates an explanation:
   \[
   c_t = C(s_t, e_t^{1}, r_{t}^1).
   \]

2. Using external feedback, the policy model $\pi_\theta$ (i.e., RAG) refines the reasoning step:
   \[
   e_{t}^{2} = \pi_\theta(s_t \mid e_{t}^{1}, c_t, r_{t}^1).
   \]

3. The PRM re-evaluates the refined step:
   \[
   r_{t}^2 = R(s_t, e_{t}^{2}).
   \]

If $r_{t}^2 > r_{t}^1$, then $\left(s_t, e_t^{1}, c_t\right)$ is labeled as a positive example with preference label $p_t = +1$. Otherwise, it is labeled as a negative example with $p_t = -1$ (cases where $r_{t}^2 = r_{t}^1$ are discarded). The collected PEM preference training dataset is denoted as:
\[
\mathcal{D}_{\text{pem}} = \left\{\left(s_t, e_{t}^{1}, c_t, r_{t}^{1}, r_{t}^{2}, p_t \right)\right\}_{i=1}^{M_{e}},
\]
where $M_{e}$ is the size of $\mathcal{D}_{\text{pem}}$.

During the training phase, since the collected preference data is binary, we employ the KTO Loss~\cite{ethayarajh2024kto} to optimize the PEM. This loss is designed for binary preference optimization and is robust to noise in the data. The KTO Loss incorporates a hyperparameter $\lambda_{U} > 1$ for negative examples, reflecting \textit{loss aversion}. In our dataset, the negative examples include the corresponding PRM scores $r_1$ and $r_2$, which can be used to dynamically adjust $\lambda_{U}$, reflecting the degree of loss aversion. Instead of assigning a uniform $\lambda_{U}$ for all negative examples as in the original KTO Loss, we introduce a dynamic $\lambda_{U}$:
\[
\lambda_{U} = \lambda_0 \cdot \exp\left(r_1 - r_2\right),
\]
where $\lambda_0$ is the base value.

\subsection{Post-Training Scaling of \mymodel}
\label{sec:post_train}

In this section, we introduce how \mymodel enhances the reasoning capabilities of RAG systems through post-training scaling. While test-time scaling can improve the reasoning performance of RAG systems to some extent, for certain weak open-source LLM-based RAG systems $\pi_{\theta}^{w}$, their inherent limitations in reasoning capabilities prevent them from solving complex multi-hop questions solely through test-time scaling. Inspired by~\cite{trung-etal-2024-reft, xie2024monte}, we propose a step-level offline reinforcement fine-tuning approach to strengthen the reasoning abilities of the RAG system.  As illustrated at the bottom of~\autoref{fig:method_2} (the retriever is omitted for simplicity), this approach comprises two stages: warm-up stage and step-level offline reinforcement stage.

\textbf{Warm-Up Stage:}
In the warm-up stage, we utilize a strong generator-based RAG system ($\pi_{\theta}^{s}$) to generate a dataset containing reasoning steps:
\[
\mathcal{D}_{w} = \{\left( q_i, \mathbf{e}_i, a_i \right) \}_{i=1}^{M_{w}},
\]
where $\mathbf{e} = [e_1, e_2, \ldots, e_T]$ and $e_t = [q_t, j_t, d_t, r_t]$. 

During fine-tuning of the weak policy $\pi_{\theta}^{w}$ using $\mathcal{D}_{w}$, the retriever-generated content $d_t$ must be masked, as it is not produced by the generator:
\[
\mathcal{L}_{w} = -\sum_{i=1}^{M_w} \sum_{t=1}^T \sum_{k=1}^{\left|e_t\right|} \mathbf{1}\left[o_{t, k} \notin d_t\right] \log \pi^{w}_\theta\left(o_{t, k} \mid q_i, o_{<t}, o_{t,<k}\right),
\]
where $o_{t, k}$ denotes the $k$-th token in sequence $e_t$, $o_{<t}$ represents all tokens generated before step $t$ (i.e., tokens in $e_1, \ldots, e_{t-1}$), and $o_{t,<k}$ denotes tokens from the first to the  position $(k-1)$ within $e_t$.

\textbf{Step-Level Offline Reinforcement Stage:}
In the step-level offline reinforcement stage, the policy model $\pi_{\theta}^{w}$ iteratively collects step-level preference data using MCTS and leverages this data to improve its reasoning capability via KTO Loss. The updated model is then used to collect new data for further policy updates.

\textit{Data Collection:}
To ensure efficient data collection and balance between positive and negative examples, we adopt the OmegaPRM-based MCTS approach~\cite{luo2024improve} introduced in~\textsection~\ref{sec:prm_train}. During rollouts, the generated nodes are scored using the PRM and PCM trained with the strong generator $\pi_{\theta}^{s}$. Reasoning steps with low reward scores are refined to gather higher-quality data.

\textit{Iterative Updates:}
In the first iteration, the policy model is initialized as $\pi_{\theta, 0}^{w}$, which is the model fine-tuned during the warm-up stage. Using the collected preference data:
\[
\mathcal{D}^{u}_{0} = \{\left(s_i, e_i, MC_i \right)\}_{i=1}^{M_{u}},
\]
where $MC_i > 0.5$ indicates desirable (positive) reasoning steps and $MC_i \leq 0.5$ indicates undesirable (negative) reasoning steps, $\mathcal{D}^{u}_{0}$ is used to update $\pi_{\theta, 0}^{w}$ via KTO Loss~\cite{ethayarajh2024kto} (with the retrieved document $d_i$ in reasoning steps $e_i$ masked). This yields the updated model $\pi_{\theta, 1}^{w}$.
In subsequent iterations, $\pi_{\theta, 1}^{w}$ is used as the policy model to generate preference data $\mathcal{D}^{u}_{1}$, which is then used to update $\pi_{\theta, 1}^{w}$ to $\pi_{\theta, 2}^{w}$. This process is repeated for $I$ iterations, progressively improving the reasoning capability of $\pi_{\theta}^{w}$. Compared to the PPO approach used in ReFT~\cite{luong2024reft}, our method sacrifices some exploration but achieves more stable updates.
Finally, $\pi_{\theta, I}^{w}$ represents the policy model after post-training scaling of \mymodel, which can be combined with test-time scaling to further enhance the reasoning capabilities of RAG systems.

\section{Experiments}
In this section, we empirically verify the effectiveness of \mymodel by addressing the following research questions:

\noindent\textbf{RQ1:} How does \mymodel improve the reasoning capabilities of RAG systems in both closed-source and open-source models?

\noindent\textbf{RQ2:} How do the components of \mymodel affect test-time scaling?

\noindent\textbf{RQ3:} How does the number of iterations during the post-training process of \mymodel affect its performance?

\noindent\textbf{RQ4:} How effective is \mymodel in aligning PEM and PRM?

\subsection{Experimental Settings}
\subsubsection{Datasets}
In this paper, we focus on leveraging \mymodel to address complex multi-step question-answering (QA) tasks. To this end, we utilize five benchmark datasets: HotpotQA~\cite{yang2018hotpotqa}, 2WikiMultiHopQA~\cite{ho2020constructing}, Musique~\cite{trivedi2022musique}, Bamboogle~\cite{press2023measuring}, and StrategyQA~\cite{geva2021did}. Wikipedia passages serve as the retrieval corpus for all datasets~\cite{jin2024flashrag}. 
Following the general experimental setup of RAG~\cite{jeong2024adaptive,kimsure,jin2024flashrag}, we sample 500 examples from the development sets of HotpotQA, 2WikiMultiHopQA, and Musique as test sets. For Bamboogle, which has only 125 examples in its test set, we include all of them as the test set. Since StrategyQA lacks dev or test sets, we sample 500 examples from its training set for testing. 

For the training data used in PRM and PCM, and for the post-training of \mymodel, we sample 200 examples from the training sets of each dataset. Using the PRM training data construction strategy described in \textsection~\ref{sec:prm_train}, we generate a total of $M_r = 167,716$ training examples. Similarly, using the PCM training data construction strategy described in \textsection~\ref{sec:pem_train}, we generate $M_e = 769$ training examples. 
For the post-training phase, the warm-up stage uses $M_w = 548$ examples, and the preference data collected during each iteration averages $M_u = 27,822$ examples.

\subsubsection{Evaluation Metrics}

During the evaluation phase, we observed that the outputs of reasoning-optimized RAG systems are typically longer compared to those generated by traditional RAG systems. Specifically, while the model accurately answers the question, it often includes extensive supplementary information. This renders exact-match metrics such as EM unsuitable for our evaluation tasks. Therefore, we adopt accuracy ($\textbf{ACC}_{R}$) as our primary evaluation metric, which determines whether the golden answer is contained within the predicted answer generated by the RAG system. To further refine our evaluation, we employ an LLM-as-Judge approach~\cite{li2024llms}, using GPT4-o~\cite{achiam2023gpt} as the evaluation model to assess whether the predicted answer is correct. This accuracy metric is referred to as $\textbf{ACC}_{L}$. The evaluation prompt is as follows:
\begin{cvbox}
Given a Question and its Golden Answer, verify whether the Predicted Answer is correct. 
The prediction is correct if it fully aligns with the meaning and key information of the Golden Answer. 
Respond with True if the prediction is correct and False otherwise.

Question: {}

Golden Answer: {}

Predicted Answer: {}
\end{cvbox}

During the Process Reward Model Training and Post-Training stages, we use $\text{ACC}_{R}$ to determine correctness in Eq.~\ref{eq:mc}, which is more efficient and better suited for collecting large amounts of training data as reward feedback.

\subsubsection{Backbone and Baseline Models.}
To verify the effectiveness of \mymodel in enhancing the reasoning capabilities of RAG systems, we selected different generators for evaluation. These include the proprietary GPT4o-mini~\cite{achiam2023gpt} for test-time scaling and the open-source LLaMA3.1-8B~\cite{dubey2024llama} (Llama-3.1-8B-Instruct) for both post-training (warm-up from GPT4o-mini) and test-time scaling.

We compared \mymodel against several baselines: 
1. \textbf{Naive Generation:} Directly generating answers using the generator without retrieval.
2. \textbf{Standard RAG:} Traditional retrieval-augmented generation systems. 
\textit{Given that \mymodel employs multi-path reasoning with CoT processes, which include adaptive retrieval and final answer generation summarized from CoTs, we further compared it with:} 3. \textbf{Branching Methods (Branching):} These execute multiple reasoning paths in parallel for a single query, including SuRe~\cite{kim2024sure} and REPLUG~\cite{shi2023replug}.
 4. \textbf{Summarization-based Methods (Summary):} LongLLMLingua~\cite{jiang2023longllmlingua}, RECOMP-abstractive~\cite{xu2024recomp}, and Selective-Context~\cite{li-etal-2023-compressing}.
5. \textbf{Adaptive Retrieval Methods (AR):} SKR~\cite{wang2023self} which adaptively retrieve based on generator's knowledge.
 6. \textbf{RAG-CoT Methods (RAG-CoT):} These integrate RAG with CoT reasoning, including Self-Ask~\cite{press2022measuring}, Iter-RetGen~\cite{shao2023enhancing}, and IRCoT~\cite{trivedi2023interleaving}.
 7. \textbf{Test-time Scaling Methods (Test-Time):} CR-Planner~\cite{li2024can}, a recently proposed approach for scaling RAG at test time.

Additionally, we compared \mymodel with LLaMA3.1-8B as the backbone against recent \textbf{Open-source Reasoning Models (Reasoning)}, such as Marco-o1-Qwen-7B~\cite{zhao2024marco} and Skywork-o1-Llama-3.1-8B~\cite{skyworkopeno12024}, which have been extensively optimized for reasoning through large-scale training in general domains and test-time scaling, both integrated into standard RAG configurations.

\begin{table*}[t]
\caption{Performance comparisons between \mymodel and the baselines. The above table shows results with GPT4-o-mini as the generator (Only Test-Time Scaling), while the below table uses LLaMA3.1-8B. The boldface indicates the best performance.}
\label{tab:main_results} 
\centering
\resizebox{0.99\linewidth}{!}{
\begin{tabular}{llcccccccccc}
\toprule 

\multirow{2}{*}{\textbf{Types}} & \multirow{2}{*}{\textbf{Models}} & \multicolumn{2}{c}{\textbf{2WikiMultiHopQA}} & \multicolumn{2}{c}{\textbf{Bamboogle}} & \multicolumn{2}{c}{\textbf{HotpotQA}} & \multicolumn{2}{c}{\textbf{Musique}} & \multicolumn{2}{c}{\textbf{StrategyQA}} \\
\cmidrule(lr){3-4} \cmidrule(lr){5-6} \cmidrule(lr){7-8} \cmidrule(lr){9-10} \cmidrule(lr){11-12}
& & $\text{ACC}_{R}$ & $\text{ACC}_{L}$ & $\text{ACC}_{R}$ & $\text{ACC}_{L}$ & $\text{ACC}_{R}$ & $\text{ACC}_{L}$ & $\text{ACC}_{R}$ & $\text{ACC}_{L}$ & $\text{ACC}_{R}$ & $\text{ACC}_{L}$ \\
\cdashline{1-12}
 \multirow{2}{*}{\textit{GPT4o-mini}} & Naive Generation & 0.348 & 0.346 & 0.240 & 0.280 & 0.324 & 0.404 & 0.134 & 0.170 & 0.724 & 0.724 \\
& Standard RAG & 0.344 & 0.292 & 0.272 & 0.328 & 0.342 & 0.450 & 0.172 & 0.188 & 0.674 & 0.674 \\

\cdashline{1-12}
\multirow{2}{*}{\textbf{Branching}} & SuRe & 0.244 & 0.264 & 0.168 & 0.208 & 0.270 & 0.380 & 0.128 & 0.146 & 0.550 & 0.576 \\
 & REPLUG & 0.296 & 0.254 & 0.224 & 0.256 & 0.350 & 0.428 & 0.132 & 0.138 & 0.654 & 0.654 \\
\cdashline{1-12}
\multirow{3}{*}{\textbf{Summary}} & LongLLMLingua & 0.324 & 0.316 & 0.248 & 0.288 & 0.358 & 0.450 & 0.150 & 0.172 & 0.722 & 0.722 \\
 & RECOMP-abstractive & 0.298 & 0.306 & 0.136 & 0.176 & 0.332 & 0.398 & 0.118 & 0.134 & 0.628 & 0.628 \\
 & Selective-Context & 0.350 & 0.290 & 0.240 & 0.288 & 0.366 & 0.442 & 0.152 & 0.172 & 0.688 & 0.688 \\
\cdashline{1-12}
\multirow{1}{*}{\textbf{Adaptive}} & SKR & 0.364 & 0.314 & 0.248 & 0.288 & 0.360 & 0.454 & 0.162 & 0.174 & 0.712 & 0.712 \\
\cdashline{1-12}
\multirow{3}{*}{\textbf{RAG-CoT}} & Self-Ask & 0.336 & 0.478 & 0.336 & 0.416 & 0.392 & 0.462 & 0.260 & 0.270 & 0.556 & 0.556 \\
 & Iter-RetGen & 0.326 & 0.270 & 0.232 & 0.256 & 0.374 & 0.456 & 0.178 & 0.188 & 0.686 & 0.686 \\
 & IRCoT & 0.492 & 0.114 & 0.272 & 0.184 & 0.434 & 0.308 & 0.192 & 0.214 & 0.406 & 0.406 \\
\cdashline{1-12}
\multirow{1}{*}{\textbf{Test-Time}} & CR-Planner & 0.520 & 0.478 & 0.488 & 0.524 & 0.404 & 0.416 & 0.272 & 0.262 & 0.744 & 0.744 \\
\cdashline{1-12}
\textbf{Ours} & \mymodel & \textbf{0.554} & \textbf{0.534} & \textbf{0.496} & \textbf{0.544} & \textbf{0.468} & \textbf{0.506} & \textbf{0.296} & \textbf{0.302} & \textbf{0.772} & \textbf{0.772} \\
\midrule
\multirow{2}{*}{\textit{LLaMA3.1-8B}} & Naive Generation & 0.326 & 0.254 & 0.144 & 0.168 & 0.208 & 0.268 & 0.068 & 0.096 & 0.672 & 0.672 \\
 & Standard RAG & 0.336 & 0.212 & 0.168 & 0.216 & 0.334 & 0.398 & 0.104 & 0.098 & 0.674 & 0.674 \\
\cdashline{1-12}
\multirow{2}{*}{\textbf{Branching}} & SuRe & 0.122 & 0.262 & 0.160 & 0.192 & 0.266 & 0.346 & 0.106 & 0.144 & 0.478 & 0.498 \\
 & REPLUG & 0.334 & 0.204 & 0.168 & 0.232 & 0.290 & 0.348 & 0.078 & 0.090 & 0.654 & 0.654 \\
\cdashline{1-12}
\multirow{3}{*}{\textbf{Summary}} & LongLLMLingua & 0.304 & 0.294 & 0.168 & 0.216 & 0.314 & 0.382 & 0.088 & 0.100 & 0.584 & 0.584 \\
 & RECOMP-abstractive & 0.324 & 0.322 & 0.104 & 0.160 & 0.318 & 0.380 & 0.112 & 0.126 & 0.628 & 0.628 \\
 & Selective-Context & 0.266 & 0.204 & 0.144 & 0.200 & 0.296 & 0.358 & 0.092 & 0.104 & 0.690 & 0.690 \\
\cdashline{1-12}
\multirow{1}{*}{\textbf{Adaptive}} & SKR & 0.336 & 0.212 & 0.176 & 0.208 & 0.300 & 0.372 & 0.100 & 0.112 & 0.662 & 0.662 \\
\cdashline{1-12}
\multirow{3}{*}{\textbf{RAG-CoT}} & Self-Ask & 0.306 & 0.322 & 0.360 & 0.432 & 0.316 & 0.408 & 0.222 & 0.226 & 0.616 & 0.616 \\
 & Iter-RetGen & 0.310 & 0.224 & 0.144 & 0.176 & 0.302 & 0.362 & 0.084 & 0.084 & 0.642 & 0.642 \\
 & IRCoT & 0.338 & 0.312 & 0.120 & 0.104 & 0.210 & 0.146 & 0.060 & 0.042 & 0.242 & 0.242 \\
\cdashline{1-12}
\multirow{1}{*}{\textbf{Test-Time}} & CR-Planer & 0.420 & 0.350 & 0.304 & 0.336 & 0.332 & 0.350 & 0.144 & 0.098 & 0.664 & 0.654 \\
\cdashline{1-12}
\multirow{2}{*}{\textbf{Reasoning}} & Marco-o1 & 0.442 & 0.184 & 0.224 & 0.200 & 0.352 & 0.348 & 0.134 & 0.104 & 0.654 & 0.504 \\
 & Skywork-o1 & 0.344 & 0.190 & 0.176 & 0.160 & 0.306 & 0.256 & 0.092 & 0.060 & 0.612 & 0.326 \\
\cdashline{1-12}
\textbf{Ours} & \mymodel & \textbf{0.470} & \textbf{0.364} & \textbf{0.438} & \textbf{0.484} & \textbf{0.424} & \textbf{0.434} & \textbf{0.244} & \textbf{0.252} & \textbf{0.724} & \textbf{0.724} \\
\bottomrule
\end{tabular}
}
\end{table*}

\subsubsection{Implementation Details}

The implementation of \mymodel and the baseline models is based on the open-source RAG framework FlashRAG~\cite{jin2024flashrag}. 
The number of samples $M$ generated per reasoning step for \mymodel at test-time is set to 3, balancing accuracy and efficiency. The maximum number of reasoning steps $T$ for Chain-of-Thought (CoT) reasoning is set to 5, where shallow nodes are defined as the first 3 reasoning steps and deep nodes are the remaining steps. 
The threshold $\tau$ for initiating the refinement phase is set to 0.5. For the lookahead search, the predefined step limit $H$ and stopping threshold $\beta$ are set to 3 and 0.05, respectively. The number $N$ in PRM training data collection is set to 5.
To ensure fairness, we configure the retrieval settings as follows: for iterative retrieval baselines and \mymodel, the number of external documents retrieved per step is set to Top 1; for single-pass retrieval baselines, the number of retrieved documents is set to 3.
The stronger generator used for collecting PRM training data is GPT4-o. The retriever utilized in all experiments is e5-base-v2~\cite{wang2022text}. For the PRM, following~\cite{li2024can} we fine-tune skywork-reward-llama-3.1-8b-v0.2~\cite{liu2024skywork} with LoRA~\cite{hulora}, which is fine-tuned from the general-purpose LLM and excels at scoring in complex scenarios. For the PEM, we fine-tune the Llama-3.2-3B-Instruct~\cite{dubey2024llama}, which is efficient and effective in generating the explanation for the policy model to refine error reasoning steps.  We run all the experiments on machines equipped with NVIDIA A6000 GPUs and 52-core Intel(R) Xeon(R) Gold 6230R CPUs at 2.10GHz.

\subsection{RQ1: Overall Performance}

\begin{table*}[t]
\caption{Ablation Study of \mymodel across different generators and datasets.}
\label{tab:ablation_results}
\centering
\resizebox{0.99\linewidth}{!}{
\begin{tabular}{llcccccccc}
\toprule
\multirow{2}{*}{\textbf{Model}} & \multirow{2}{*}{\textbf{Ablation}} & \multicolumn{2}{c}{\textbf{2WikiMultiHopQA}} & \multicolumn{2}{c}{\textbf{Bamboogle}} & \multicolumn{2}{c}{\textbf{HotpotQA}} & \multicolumn{2}{c}{\textbf{Musique}} \\
\cmidrule(lr){3-4} \cmidrule(lr){5-6} \cmidrule(lr){7-8} \cmidrule(lr){9-10}
& & $\text{ACC}_{R}$ & $\text{ACC}_{L}$ & $\text{ACC}_{R}$ & $\text{ACC}_{L}$ & $\text{ACC}_{R}$ & $\text{ACC}_{L}$ & $\text{ACC}_{R}$ & $\text{ACC}_{L}$ \\
\midrule
\multirow{6}{*}{\textbf{GPT4o-mini}} 
& w/o Refinement & 0.522 & 0.466 & 0.474 & 0.522 & 0.424 & 0.456 & 0.282 & 0.276 \\
& w/o PEM & 0.532 & 0.484 & 0.486 & 0.532 & 0.426 & 0.462 & 0.284 & 0.286 \\
& w/o TD-Lookahead & 0.524 & 0.490 & 0.488 & 0.540 & 0.458 & 0.494 & 0.290 & 0.294 \\
& w/o Beam Search & 0.526 & 0.492 & 0.482 & 0.522 & 0.442 & 0.474 & 0.278 & 0.272 \\
& w/o PRM Data & 0.536 & 0.476 & 0.474 & 0.534 & 0.464 & 0.504 & 0.288 & 0.290 \\
& \mymodel & \textbf{0.554} & \textbf{0.534} & \textbf{0.496} & \textbf{0.544} & \textbf{0.468} & \textbf{0.506} & \textbf{0.296} & \textbf{0.302} \\
\midrule
\multirow{6}{*}{\textbf{Llama-3.1-8B}} 
& w/o Refinement & 0.444 & 0.334 & 0.418 & 0.440 & 0.402 & 0.424 & 0.230 & 0.238 \\
& w/o PEM & 0.450 & 0.340 & 0.420 & 0.446 & 0.406 & 0.416 & 0.234 & 0.218 \\
& w/o TD-Lookahead & 0.462 & 0.352 & 0.428 & 0.454 & 0.414 & 0.438 & 0.222 & 0.242 \\
& w/o Beam Search & 0.452 & 0.346 & 0.424 & 0.448 & 0.416 & 0.420 & 0.236 & 0.246 \\
& w/o PRM Data & 0.466 & 0.350 & 0.416 & 0.458 & 0.406 & 0.400 & 0.238 & 0.232 \\
& \mymodel & \textbf{0.470} & \textbf{0.364} & \textbf{0.438} & \textbf{0.484} & \textbf{0.424} & \textbf{0.434} & \textbf{0.244} & \textbf{0.252} \\
\bottomrule
\end{tabular}
}
\end{table*}

\autoref{tab:main_results} presents the experimental results of applying \mymodel to RAG systems with two different generators: the proprietary GPT4o-mini and the open-source LLaMA3.1-8B, across five multi-step QA datasets. For the RAG system with GPT4o-mini as the generator, where fine-tuning is not feasible, we applied only the Test-Time Scaling component of \mymodel. Based on the results in \autoref{tab:main_results}, we observed the following key findings:
(1) Compared to baseline models, \mymodel significantly improves the reasoning capabilities of RAG systems in both closed-source and open-source setups, demonstrating the generalizability of the \mymodel framework in enhancing RAG systems' reasoning abilities.
(2) \mymodel outperforms Branching methods, indicating that multi-path exploration through Chain-of-Thought (CoT) reasoning is better suited for complex multi-step QA tasks than probability integration in REPLUG or Best-of-K strategies in SuRe.
(3) \mymodel surpasses summarization-based methods, suggesting that conducting CoT reasoning followed by summarization is superior to directly compressing and summarizing external document knowledge for multi-step reasoning tasks.
(4) \mymodel outperforms adaptive retrieval methods, showing that allowing the generator to dynamically decide whether to retrieve in the CoT process can further unlock the model's reasoning potential and improve its ability to answer complex questions.
(5) \mymodel exceeds the performance of RAG-CoT methods, demonstrating that our approach, which leverages Post-Training and Test-Time Scaling, more effectively enhances reasoning capabilities compared to directly combining RAG and CoT reasoning.
(6) \mymodel outperforms CR-Planner, validating that our proposed Trustworthy Process Rewarding mechanism produces superior reasoning paths for RAG systems, thereby improving their ability to handle complex multi-step reasoning problems.
(7) \mymodel surpasses models extensively optimized for reasoning through large-scale training on general domains and test-time scaling, such as Skywork-o1 and Marco-o1. This result indicates that models optimized for general tasks are less effective in RAG-specific reasoning scenarios compared to our framework, further highlighting the effectiveness of \mymodel in enhancing the reasoning capabilities of RAG systems.

\begin{figure}[htbp]
    \centering
    \begin{subfigure}{0.49\linewidth}
        \centering
        \includegraphics[width=\textwidth]{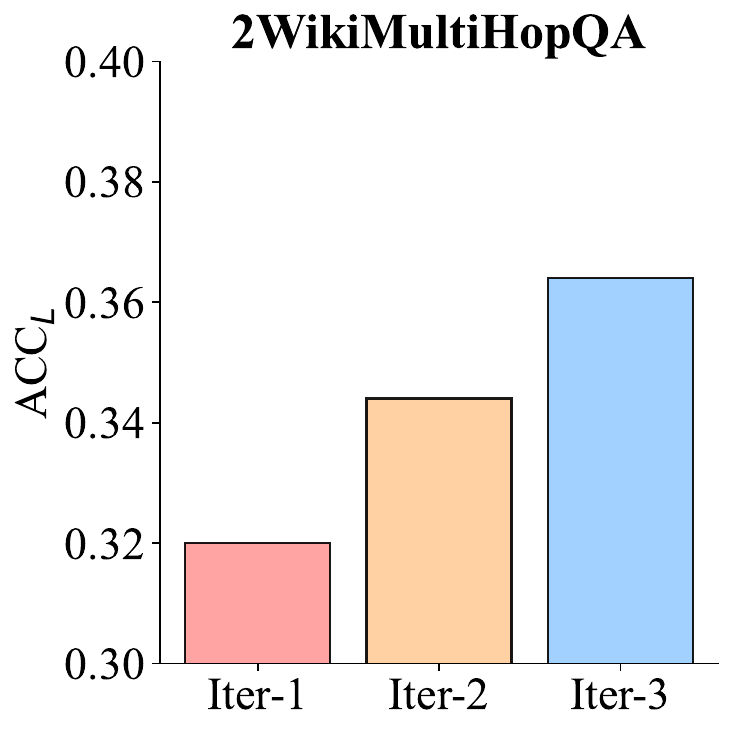} 
        \label{subfig:image1}
    \end{subfigure}
    \hfill
    \begin{subfigure}{0.49\linewidth}
        \centering
        \includegraphics[width=\textwidth]{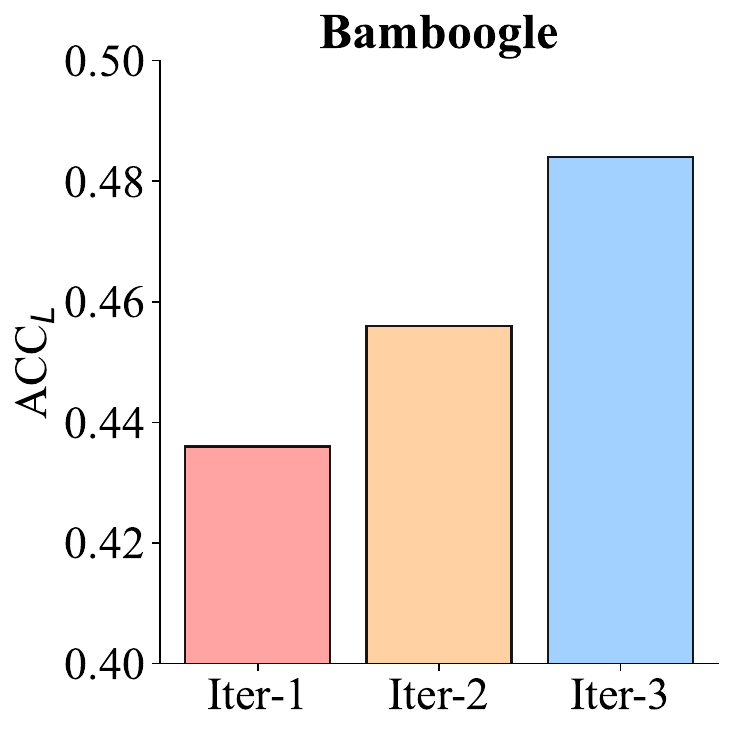} 
        \label{subfig:image2}
    \end{subfigure}

    \vspace{0.5em} 

    \begin{subfigure}{0.49\linewidth}
        \centering
        \includegraphics[width=\textwidth]{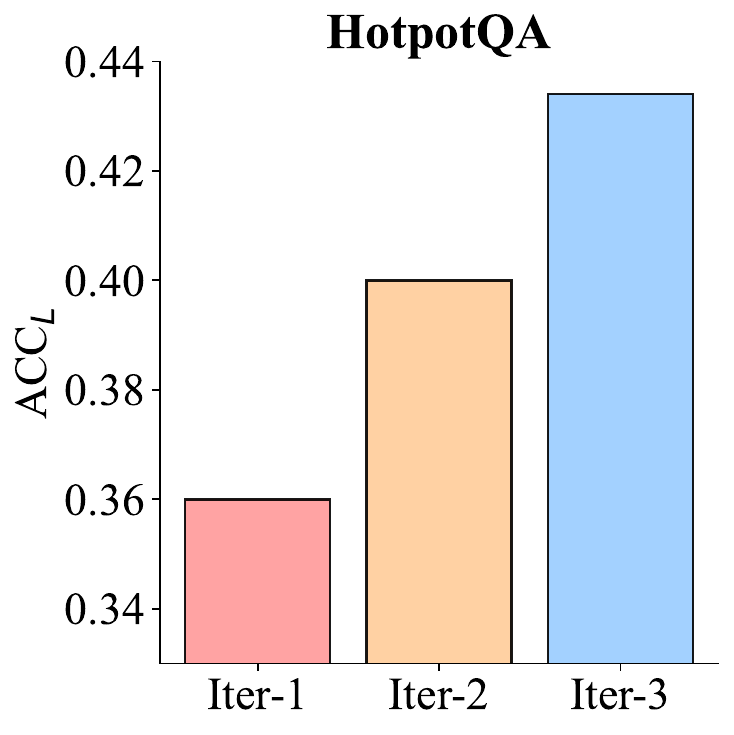} 
        \label{subfig:image3}
    \end{subfigure}
    \hfill
    \begin{subfigure}{0.49\linewidth}
        \centering
        \includegraphics[width=\textwidth]{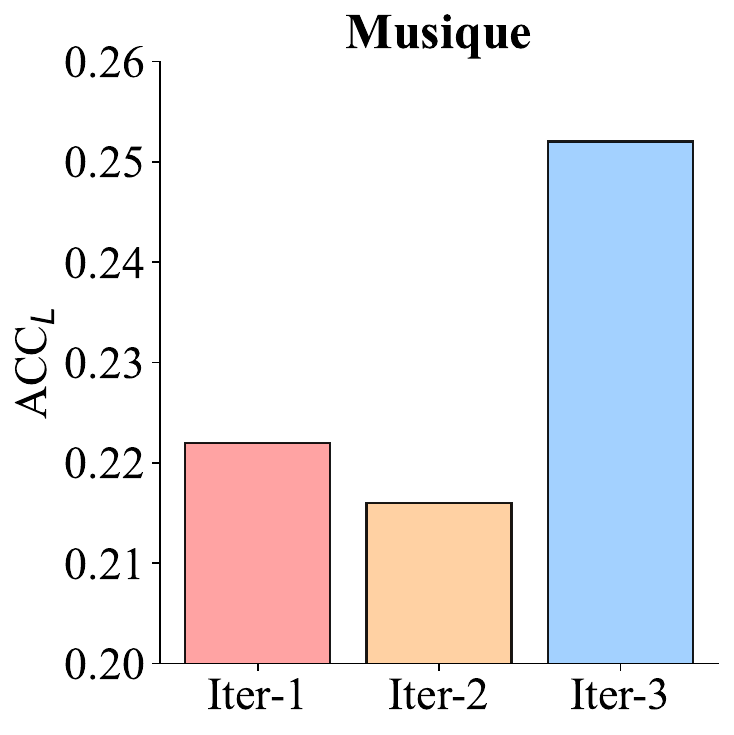} 
        \label{subfig:image4}
    \end{subfigure}

    \caption{The impact of Post-Training Scaling iterations on \mymodel using LLaMA-3.1-8B as the generator.} 
    \label{fig:iter_results}
\end{figure}

\subsection{RQ2: Ablation Study of \mymodel}
In this section, we conduct an ablation study to analyze the impact of different components of \mymodel on the test-time scaling performance of RAG systems. Specifically, we evaluate the following configurations:
(1) \textbf{w/o Refinement}: Removing the refinement phase to analyze its effect on the reasoning process of \mymodel.
(2) \textbf{w/o PEM}: Replacing the Process Explanation Model (PEM) with the process reward score directly provided by the PRM during the refinement phase, to evaluate the importance of PEM-generated explanations for refinement.
(3) \textbf{w/o TD-Lookahead}: Removing the TD-based lookahead search to validate its role in mitigating early-step bias in the PRM.
(4) \textbf{w/o PRM Data}: Training the PRM using data collected with traditional Monte Carlo methods instead of the unbiased data collection strategy proposed in \mymodel, to analyze the quality of the PRM trained with our data collection method.
(5) \textbf{w/o Beam Search}: Disabling beam search by setting $M = 1$, resulting in only a single reasoning path being sampled to generate the CoT.

The experimental results, presented in \autoref{tab:main_results}, demonstrate that removing any of these components negatively impacts the overall performance of \mymodel. This highlights the importance of each component in enhancing the reasoning capabilities of the RAG system. Moreover, these results validate that the unbiased PRM training data collection strategy designed to address the untrustworthy challenges of process reward models enables the training of a more reliable PRM, which provides accurate reward scores for reasoning steps.
Additionally, the combination of a more accurate PRM with the TD-based lookahead search enhances the feedback provided during the refinement stage. By leveraging explanations generated by the PEM during the refinement phase, \mymodel achieves better reasoning step improvements compared to using PRM scores alone.

\subsection{RQ3: Post-training iterations analysis.}
In this section, we analyze the impact of the number of iterations in the Step-Level Offline Reinforcement Stage during the post-training scaling of \mymodel on the reasoning capabilities of RAG systems. In this experiment, we used LLaMA-3.1-8B as the generator and conducted three iterations of Offline Reinforcement, testing the system on four multi-step reasoning datasets.

The experimental results, presented in \autoref{fig:iter_results}, demonstrate that the performance of the RAG system on multi-step reasoning datasets improves significantly as the number of Offline Reinforcement iterations increases. Additionally, the results show that our algorithm achieves stable performance improvements across iterations, validating that the proposed Step-Level Offline Reinforcement method provides effective and consistent updates.
Due to resource constraints, we did not verify the scalability of our approach on larger datasets or with additional iterations. However, based on the current experimental results, we observe a promising scaling property, suggesting the potential for even greater improvements under resource-abundant conditions.

\begin{figure}[tbp]
    \centering
    \begin{subfigure}{\linewidth}
        \centering
        \includegraphics[width=\textwidth]{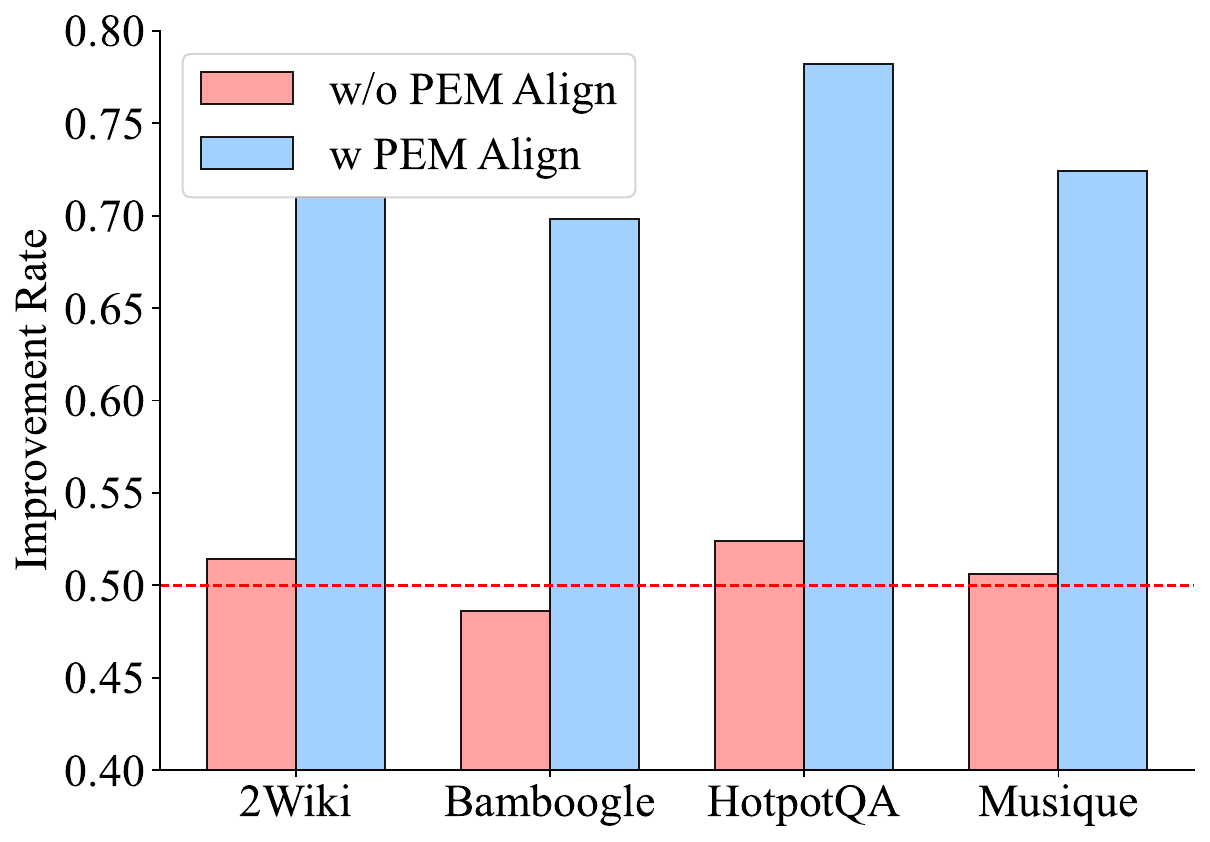} 
        \subcaption{The impact of alignment between PEM and PRM on the improvement rate of process reward scores.}
    \end{subfigure}

    \vspace{0.5em} 

    \begin{subfigure}{\linewidth}
        \centering
        \includegraphics[width=\textwidth]{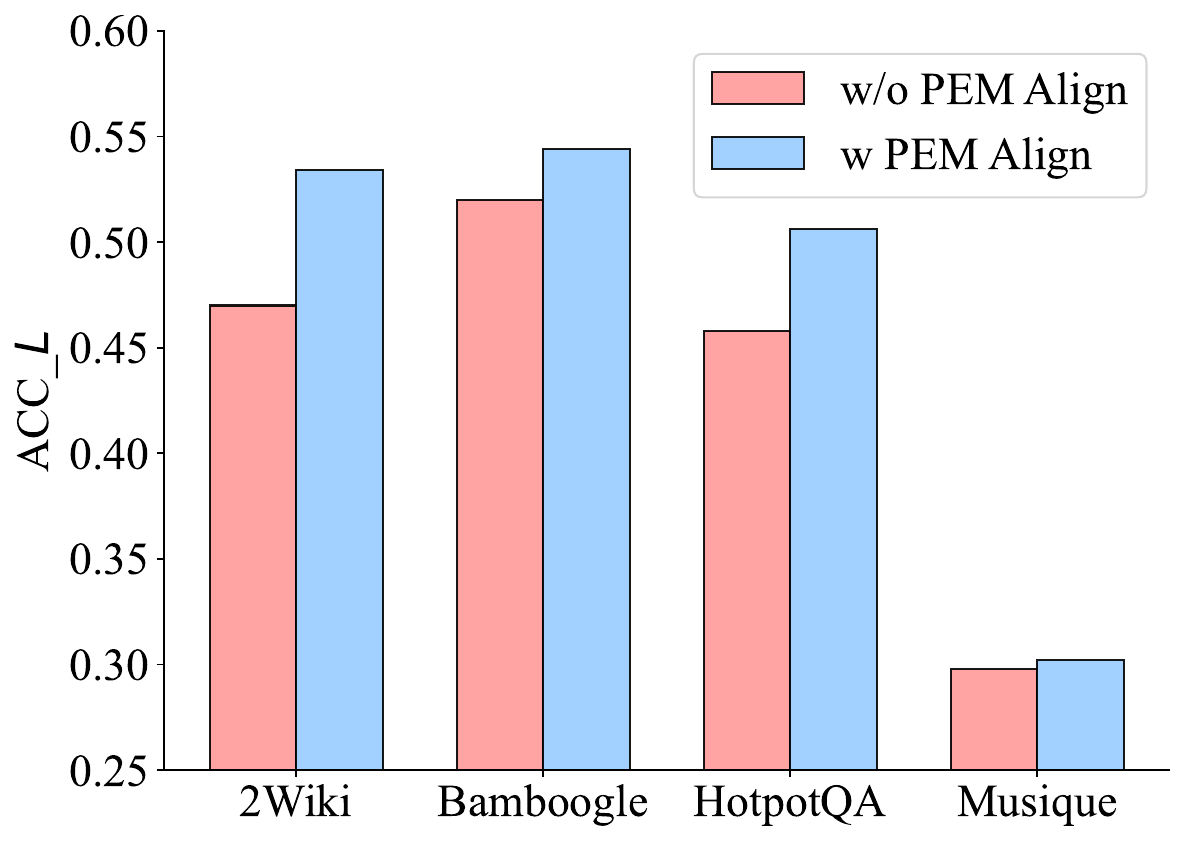} 
        \subcaption{The impact of PEM on \mymodel's final performance before and after alignment with PRM.}
    \end{subfigure}

    \caption{The impact of aligning PEM and PRM on \mymodel's overall performance with GPT4-o-mini as generator.} 
    \label{fig:PEM}
\end{figure}

\subsection{RQ4: The effective of RARTPR in aligning PEM and PRM.}
To evaluate the effectiveness of the alignment strategy for PEM and PRM proposed in RARTPR, we first calculated the improvement rate of process reward scores for reasoning steps with low initial scores after refinement using explanations generated by PEM, both before and after alignment. As shown in \autoref{fig:PEM}(a), before aligning PEM with PRM (\textit{w/o PEM Align}), the improvement rate achieved using explanations from an off-the-shelf LLM-based PEM was only around 50\%. This result indicates that PEM struggles to produce accurate explanations aligned with PRM scores, making it difficult for the RAG system to leverage these explanations to refine reasoning steps and improve PRM scores. 
In contrast, after aligning PEM and PRM (\textit{w PEM Align}), we observed a significant increase in the improvement rate, validating the effectiveness of the alignment strategy for enhancing the refinement process and improving reasoning quality.
Furthermore, as shown in \autoref{fig:PEM}(b), we directly compared the accuracy of RARTPR in solving complex multi-hop queries before and after aligning PEM with PRM. The results demonstrate consistent improvements across multiple datasets after alignment, further confirming the effectiveness of the proposed alignment strategy for PEM and PRM in RARTPR.

We further present a case study of explanations generated by the aligned PEM to demonstrate the effectiveness of PEM under process supervision:
\begin{cvbox}
Based on the provided information, I've evaluated the Actor model's current reasoning step and identified the following errors:

\textbf{Question Decomposition Errors}

- 1.1: The sub-question "Who was the father of John V, Prince of Anhalt-Zerbst?" is a valid sub-question, but the subsequent sub-question "Follow up: Who was John V, Prince Of Anhalt-Zerbst's father?" is not directly related to the original question. This can be improved by rephrasing the sub-question to better align with the original query.

\textbf{Retrieval Errors}

- 2.1: The retrieved documents contain noise or erroneous information, as the intermediate answer states that the retrieved document does not provide information relevant to the death of John IV, Prince of Anhalt-Zerbst, despite the search result being "Yes".

- 2.2: The retrieved documents are inadequate or irrelevant, as the intermediate answer fails to provide a clear answer to the original question.

\textbf{Answer Generation Errors}

- 3.1: The answer "John V, Prince of Anhalt-Zerbst died in 1454" is irrelevant to the query, as the original question asks for the death date of John IV, Prince of Anhalt-Zerbst.

- 3.2: The answer is incomplete, as the original question asks for the death date of John IV, Prince of Anhalt-Zerbst.

\textbf{Suggestions for improvement:}

1. Refine the sub-question decomposition to ensure that subsequent sub-questions are directly related to the original query.
2. Improve the retrieval process to ensure that relevant and accurate information is retrieved.
3. Verify the accuracy of the answer generation process to provide relevant and correct information.
4. Consider adding additional checks to ensure that the answer is complete and relevant to the original question.
\end{cvbox}

\section{Conclusion}

We propose \textbf{\mymodel}, a framework that enhances the multi-step reasoning capabilities of RAG systems through both post-training and test-time scaling. \mymodel integrates Trustworthy Process Rewarding, which combines a Process Reward Model for accurate scoring and a Process Explanation Model for explanation-based refinements. 
During post-training, step-level offline reinforcement fine-tuning with MCTS generates high-quality preference data to optimize the generator. \mymodel addresses key reasoning challenges, including misalignment between PEM and PRM, bias in PRM training data, and early-step bias in PRM scores, through off-policy preference learning, balanced annotation strategies, and a temporal-difference-based look-ahead search. Experiments on multi-step reasoning benchmarks show that \mymodel outperforms existing methods, demonstrating its effectiveness in enhancing RAG systems for knowledge-intensive tasks.


\bibliographystyle{ACM-Reference-Format}
\bibliography{sample-base}


\begin{thebibliography}{49}


\ifx \showCODEN    \undefined \def \showCODEN     #1{\unskip}     \fi
\ifx \showDOI      \undefined \def \showDOI       #1{#1}\fi
\ifx \showISBNx    \undefined \def \showISBNx     #1{\unskip}     \fi
\ifx \showISBNxiii \undefined \def \showISBNxiii  #1{\unskip}     \fi
\ifx \showISSN     \undefined \def \showISSN      #1{\unskip}     \fi
\ifx \showLCCN     \undefined \def \showLCCN      #1{\unskip}     \fi
\ifx \shownote     \undefined \def \shownote      #1{#1}          \fi
\ifx \showarticletitle \undefined \def \showarticletitle #1{#1}   \fi
\ifx \showURL      \undefined \def \showURL       {\relax}        \fi
\providecommand\bibfield[2]{#2}
\providecommand\bibinfo[2]{#2}
\providecommand\natexlab[1]{#1}
\providecommand\showeprint[2][]{arXiv:#2}

\bibitem[Achiam et~al\mbox{.}(2023)]%
        {achiam2023gpt}
\bibfield{author}{\bibinfo{person}{Josh Achiam}, \bibinfo{person}{Steven Adler}, \bibinfo{person}{Sandhini Agarwal}, \bibinfo{person}{Lama Ahmad}, \bibinfo{person}{Ilge Akkaya}, \bibinfo{person}{Florencia~Leoni Aleman}, \bibinfo{person}{Diogo Almeida}, \bibinfo{person}{Janko Altenschmidt}, \bibinfo{person}{Sam Altman}, \bibinfo{person}{Shyamal Anadkat}, {et~al\mbox{.}}} \bibinfo{year}{2023}\natexlab{}.
\newblock \showarticletitle{Gpt-4 technical report}.
\newblock \bibinfo{journal}{\emph{arXiv preprint arXiv:2303.08774}} (\bibinfo{year}{2023}).
\newblock


\bibitem[Browne et~al\mbox{.}(2012)]%
        {browne2012survey}
\bibfield{author}{\bibinfo{person}{Cameron~B Browne}, \bibinfo{person}{Edward Powley}, \bibinfo{person}{Daniel Whitehouse}, \bibinfo{person}{Simon~M Lucas}, \bibinfo{person}{Peter~I Cowling}, \bibinfo{person}{Philipp Rohlfshagen}, \bibinfo{person}{Stephen Tavener}, \bibinfo{person}{Diego Perez}, \bibinfo{person}{Spyridon Samothrakis}, {and} \bibinfo{person}{Simon Colton}.} \bibinfo{year}{2012}\natexlab{}.
\newblock \showarticletitle{A survey of monte carlo tree search methods}.
\newblock \bibinfo{journal}{\emph{IEEE Transactions on Computational Intelligence and AI in games}} \bibinfo{volume}{4}, \bibinfo{number}{1} (\bibinfo{year}{2012}), \bibinfo{pages}{1--43}.
\newblock


\bibitem[Dong et~al\mbox{.}(2024)]%
        {dong2024progressive}
\bibfield{author}{\bibinfo{person}{Guanting Dong}, \bibinfo{person}{Chenghao Zhang}, \bibinfo{person}{Mengjie Deng}, \bibinfo{person}{Yutao Zhu}, \bibinfo{person}{Zhicheng Dou}, {and} \bibinfo{person}{Ji-Rong Wen}.} \bibinfo{year}{2024}\natexlab{}.
\newblock \showarticletitle{Progressive Multimodal Reasoning via Active Retrieval}.
\newblock \bibinfo{journal}{\emph{arXiv preprint arXiv:2412.14835}} (\bibinfo{year}{2024}).
\newblock


\bibitem[Dubey et~al\mbox{.}(2024)]%
        {dubey2024llama}
\bibfield{author}{\bibinfo{person}{Abhimanyu Dubey}, \bibinfo{person}{Abhinav Jauhri}, \bibinfo{person}{Abhinav Pandey}, \bibinfo{person}{Abhishek Kadian}, \bibinfo{person}{Ahmad Al-Dahle}, \bibinfo{person}{Aiesha Letman}, \bibinfo{person}{Akhil Mathur}, \bibinfo{person}{Alan Schelten}, \bibinfo{person}{Amy Yang}, \bibinfo{person}{Angela Fan}, {et~al\mbox{.}}} \bibinfo{year}{2024}\natexlab{}.
\newblock \showarticletitle{The llama 3 herd of models}.
\newblock \bibinfo{journal}{\emph{arXiv preprint arXiv:2407.21783}} (\bibinfo{year}{2024}).
\newblock


\bibitem[Ethayarajh et~al\mbox{.}(2024)]%
        {ethayarajh2024kto}
\bibfield{author}{\bibinfo{person}{Kawin Ethayarajh}, \bibinfo{person}{Winnie Xu}, \bibinfo{person}{Niklas Muennighoff}, \bibinfo{person}{Dan Jurafsky}, {and} \bibinfo{person}{Douwe Kiela}.} \bibinfo{year}{2024}\natexlab{}.
\newblock \showarticletitle{Kto: Model alignment as prospect theoretic optimization}.
\newblock \bibinfo{journal}{\emph{arXiv preprint arXiv:2402.01306}} (\bibinfo{year}{2024}).
\newblock


\bibitem[Fan et~al\mbox{.}(2024)]%
        {ragsurvey}
\bibfield{author}{\bibinfo{person}{Wenqi Fan}, \bibinfo{person}{Yujuan Ding}, \bibinfo{person}{Liangbo Ning}, \bibinfo{person}{Shijie Wang}, \bibinfo{person}{Hengyun Li}, \bibinfo{person}{Dawei Yin}, \bibinfo{person}{Tat-Seng Chua}, {and} \bibinfo{person}{Qing Li}.} \bibinfo{year}{2024}\natexlab{}.
\newblock \showarticletitle{A Survey on RAG Meeting LLMs: Towards Retrieval-Augmented Large Language Models}. In \bibinfo{booktitle}{\emph{Proceedings of the 30th ACM SIGKDD Conference on Knowledge Discovery and Data Mining}} (Barcelona, Spain) \emph{(\bibinfo{series}{KDD '24})}. \bibinfo{publisher}{Association for Computing Machinery}, \bibinfo{address}{New York, NY, USA}, \bibinfo{pages}{6491–6501}.
\newblock
\showISBNx{9798400704901}
\urldef\tempurl%
\url{https://doi.org/10.1145/3637528.3671470}
\showDOI{\tempurl}


\bibitem[Geva et~al\mbox{.}(2021)]%
        {geva2021did}
\bibfield{author}{\bibinfo{person}{Mor Geva}, \bibinfo{person}{Daniel Khashabi}, \bibinfo{person}{Elad Segal}, \bibinfo{person}{Tushar Khot}, \bibinfo{person}{Dan Roth}, {and} \bibinfo{person}{Jonathan Berant}.} \bibinfo{year}{2021}\natexlab{}.
\newblock \showarticletitle{Did aristotle use a laptop? a question answering benchmark with implicit reasoning strategies}.
\newblock \bibinfo{journal}{\emph{Transactions of the Association for Computational Linguistics}}  \bibinfo{volume}{9} (\bibinfo{year}{2021}), \bibinfo{pages}{346--361}.
\newblock


\bibitem[Ho et~al\mbox{.}(2020)]%
        {ho2020constructing}
\bibfield{author}{\bibinfo{person}{Xanh Ho}, \bibinfo{person}{Anh-Khoa~Duong Nguyen}, \bibinfo{person}{Saku Sugawara}, {and} \bibinfo{person}{Akiko Aizawa}.} \bibinfo{year}{2020}\natexlab{}.
\newblock \showarticletitle{Constructing A Multi-hop QA Dataset for Comprehensive Evaluation of Reasoning Steps}. In \bibinfo{booktitle}{\emph{Proceedings of the 28th International Conference on Computational Linguistics}}. \bibinfo{pages}{6609--6625}.
\newblock


\bibitem[Hu et~al\mbox{.}({[n.\,d.]})]%
        {hulora}
\bibfield{author}{\bibinfo{person}{Edward~J Hu}, \bibinfo{person}{Phillip Wallis}, \bibinfo{person}{Zeyuan Allen-Zhu}, \bibinfo{person}{Yuanzhi Li}, \bibinfo{person}{Shean Wang}, \bibinfo{person}{Lu Wang}, \bibinfo{person}{Weizhu Chen}, {et~al\mbox{.}}} \bibinfo{year}{[n.\,d.]}\natexlab{}.
\newblock \showarticletitle{LoRA: Low-Rank Adaptation of Large Language Models}. In \bibinfo{booktitle}{\emph{International Conference on Learning Representations}}.
\newblock


\bibitem[Huang et~al\mbox{.}(2024)]%
        {huang2024large}
\bibfield{author}{\bibinfo{person}{Jie Huang}, \bibinfo{person}{Xinyun Chen}, \bibinfo{person}{Swaroop Mishra}, \bibinfo{person}{Huaixiu~Steven Zheng}, \bibinfo{person}{Adams~Wei Yu}, \bibinfo{person}{Xinying Song}, {and} \bibinfo{person}{Denny Zhou}.} \bibinfo{year}{2024}\natexlab{}.
\newblock \showarticletitle{Large Language Models Cannot Self-Correct Reasoning Yet}. In \bibinfo{booktitle}{\emph{The Twelfth International Conference on Learning Representations}}.
\newblock
\urldef\tempurl%
\url{https://openreview.net/forum?id=IkmD3fKBPQ}
\showURL{%
\tempurl}


\bibitem[Jeong et~al\mbox{.}(2024a)]%
        {adaptiverag}
\bibfield{author}{\bibinfo{person}{Soyeong Jeong}, \bibinfo{person}{Jinheon Baek}, \bibinfo{person}{Sukmin Cho}, \bibinfo{person}{Sung~Ju Hwang}, {and} \bibinfo{person}{Jong Park}.} \bibinfo{year}{2024}\natexlab{a}.
\newblock \showarticletitle{Adaptive-{RAG}: Learning to Adapt Retrieval-Augmented Large Language Models through Question Complexity}. In \bibinfo{booktitle}{\emph{Proceedings of the 2024 Conference of the North American Chapter of the Association for Computational Linguistics: Human Language Technologies (Volume 1: Long Papers)}}, \bibfield{editor}{\bibinfo{person}{Kevin Duh}, \bibinfo{person}{Helena Gomez}, {and} \bibinfo{person}{Steven Bethard}} (Eds.). \bibinfo{publisher}{Association for Computational Linguistics}, \bibinfo{address}{Mexico City, Mexico}, \bibinfo{pages}{7036--7050}.
\newblock
\urldef\tempurl%
\url{https://doi.org/10.18653/v1/2024.naacl-long.389}
\showDOI{\tempurl}


\bibitem[Jeong et~al\mbox{.}(2024b)]%
        {jeong2024adaptive}
\bibfield{author}{\bibinfo{person}{Soyeong Jeong}, \bibinfo{person}{Jinheon Baek}, \bibinfo{person}{Sukmin Cho}, \bibinfo{person}{Sung~Ju Hwang}, {and} \bibinfo{person}{Jong~C Park}.} \bibinfo{year}{2024}\natexlab{b}.
\newblock \showarticletitle{Adaptive-RAG: Learning to Adapt Retrieval-Augmented Large Language Models through Question Complexity}. In \bibinfo{booktitle}{\emph{Proceedings of the 2024 Conference of the North American Chapter of the Association for Computational Linguistics: Human Language Technologies (Volume 1: Long Papers)}}. \bibinfo{pages}{7029--7043}.
\newblock


\bibitem[Jiang et~al\mbox{.}(2023)]%
        {jiang2023longllmlingua}
\bibfield{author}{\bibinfo{person}{Huiqiang Jiang}, \bibinfo{person}{Qianhui Wu}, \bibinfo{person}{Xufang Luo}, \bibinfo{person}{Dongsheng Li}, \bibinfo{person}{Chin-Yew Lin}, \bibinfo{person}{Yuqing Yang}, {and} \bibinfo{person}{Lili Qiu}.} \bibinfo{year}{2023}\natexlab{}.
\newblock \showarticletitle{Longllmlingua: Accelerating and enhancing llms in long context scenarios via prompt compression}.
\newblock \bibinfo{journal}{\emph{arXiv preprint arXiv:2310.06839}} (\bibinfo{year}{2023}).
\newblock


\bibitem[Jin et~al\mbox{.}(2024)]%
        {jin2024flashrag}
\bibfield{author}{\bibinfo{person}{Jiajie Jin}, \bibinfo{person}{Yutao Zhu}, \bibinfo{person}{Xinyu Yang}, \bibinfo{person}{Chenghao Zhang}, {and} \bibinfo{person}{Zhicheng Dou}.} \bibinfo{year}{2024}\natexlab{}.
\newblock \showarticletitle{FlashRAG: A Modular Toolkit for Efficient Retrieval-Augmented Generation Research}.
\newblock \bibinfo{journal}{\emph{arXiv preprint arXiv:2405.13576}} (\bibinfo{year}{2024}).
\newblock


\bibitem[Kim et~al\mbox{.}({[n.\,d.]})]%
        {kimsure}
\bibfield{author}{\bibinfo{person}{Jaehyung Kim}, \bibinfo{person}{Jaehyun Nam}, \bibinfo{person}{Sangwoo Mo}, \bibinfo{person}{Jongjin Park}, \bibinfo{person}{Sang-Woo Lee}, \bibinfo{person}{Minjoon Seo}, \bibinfo{person}{Jung-Woo Ha}, {and} \bibinfo{person}{Jinwoo Shin}.} \bibinfo{year}{[n.\,d.]}\natexlab{}.
\newblock \showarticletitle{SuRe: Summarizing Retrievals using Answer Candidates for Open-domain QA of LLMs}. In \bibinfo{booktitle}{\emph{The Twelfth International Conference on Learning Representations}}.
\newblock


\bibitem[Kim et~al\mbox{.}(2024)]%
        {kim2024sure}
\bibfield{author}{\bibinfo{person}{Jaehyung Kim}, \bibinfo{person}{Jaehyun Nam}, \bibinfo{person}{Sangwoo Mo}, \bibinfo{person}{Jongjin Park}, \bibinfo{person}{Sang-Woo Lee}, \bibinfo{person}{Minjoon Seo}, \bibinfo{person}{Jung-Woo Ha}, {and} \bibinfo{person}{Jinwoo Shin}.} \bibinfo{year}{2024}\natexlab{}.
\newblock \showarticletitle{SuRe: Summarizing Retrievals using Answer Candidates for Open-domain {QA} of {LLM}s}. In \bibinfo{booktitle}{\emph{The Twelfth International Conference on Learning Representations}}.
\newblock
\urldef\tempurl%
\url{https://openreview.net/forum?id=w4DW6qkRmt}
\showURL{%
\tempurl}


\bibitem[Li et~al\mbox{.}(2024a)]%
        {li2024llms}
\bibfield{author}{\bibinfo{person}{Haitao Li}, \bibinfo{person}{Qian Dong}, \bibinfo{person}{Junjie Chen}, \bibinfo{person}{Huixue Su}, \bibinfo{person}{Yujia Zhou}, \bibinfo{person}{Qingyao Ai}, \bibinfo{person}{Ziyi Ye}, {and} \bibinfo{person}{Yiqun Liu}.} \bibinfo{year}{2024}\natexlab{a}.
\newblock \showarticletitle{LLMs-as-Judges: A Comprehensive Survey on LLM-based Evaluation Methods}.
\newblock \bibinfo{journal}{\emph{arXiv preprint arXiv:2412.05579}} (\bibinfo{year}{2024}).
\newblock


\bibitem[Li et~al\mbox{.}(2024b)]%
        {li2024can}
\bibfield{author}{\bibinfo{person}{Xingxuan Li}, \bibinfo{person}{Weiwen Xu}, \bibinfo{person}{Ruochen Zhao}, \bibinfo{person}{Fangkai Jiao}, \bibinfo{person}{Shafiq Joty}, {and} \bibinfo{person}{Lidong Bing}.} \bibinfo{year}{2024}\natexlab{b}.
\newblock \showarticletitle{Can We Further Elicit Reasoning in LLMs? Critic-Guided Planning with Retrieval-Augmentation for Solving Challenging Tasks}.
\newblock \bibinfo{journal}{\emph{arXiv preprint arXiv:2410.01428}} (\bibinfo{year}{2024}).
\newblock


\bibitem[Li et~al\mbox{.}(2023)]%
        {li-etal-2023-compressing}
\bibfield{author}{\bibinfo{person}{Yucheng Li}, \bibinfo{person}{Bo Dong}, \bibinfo{person}{Frank Guerin}, {and} \bibinfo{person}{Chenghua Lin}.} \bibinfo{year}{2023}\natexlab{}.
\newblock \showarticletitle{Compressing Context to Enhance Inference Efficiency of Large Language Models}. In \bibinfo{booktitle}{\emph{Proceedings of the 2023 Conference on Empirical Methods in Natural Language Processing}}, \bibfield{editor}{\bibinfo{person}{Houda Bouamor}, \bibinfo{person}{Juan Pino}, {and} \bibinfo{person}{Kalika Bali}} (Eds.). \bibinfo{publisher}{Association for Computational Linguistics}, \bibinfo{address}{Singapore}, \bibinfo{pages}{6342--6353}.
\newblock
\urldef\tempurl%
\url{https://doi.org/10.18653/v1/2023.emnlp-main.391}
\showDOI{\tempurl}


\bibitem[Liu et~al\mbox{.}(2024)]%
        {liu2024skywork}
\bibfield{author}{\bibinfo{person}{Chris~Yuhao Liu}, \bibinfo{person}{Liang Zeng}, \bibinfo{person}{Jiacai Liu}, \bibinfo{person}{Rui Yan}, \bibinfo{person}{Jujie He}, \bibinfo{person}{Chaojie Wang}, \bibinfo{person}{Shuicheng Yan}, \bibinfo{person}{Yang Liu}, {and} \bibinfo{person}{Yahui Zhou}.} \bibinfo{year}{2024}\natexlab{}.
\newblock \showarticletitle{Skywork-Reward: Bag of Tricks for Reward Modeling in LLMs}.
\newblock \bibinfo{journal}{\emph{arXiv preprint arXiv:2410.18451}} (\bibinfo{year}{2024}).
\newblock


\bibitem[Luo et~al\mbox{.}(2024)]%
        {luo2024improve}
\bibfield{author}{\bibinfo{person}{Liangchen Luo}, \bibinfo{person}{Yinxiao Liu}, \bibinfo{person}{Rosanne Liu}, \bibinfo{person}{Samrat Phatale}, \bibinfo{person}{Harsh Lara}, \bibinfo{person}{Yunxuan Li}, \bibinfo{person}{Lei Shu}, \bibinfo{person}{Yun Zhu}, \bibinfo{person}{Lei Meng}, \bibinfo{person}{Jiao Sun}, {et~al\mbox{.}}} \bibinfo{year}{2024}\natexlab{}.
\newblock \showarticletitle{Improve Mathematical Reasoning in Language Models by Automated Process Supervision}.
\newblock \bibinfo{journal}{\emph{arXiv preprint arXiv:2406.06592}} (\bibinfo{year}{2024}).
\newblock


\bibitem[Luong et~al\mbox{.}(2024)]%
        {luong2024reft}
\bibfield{author}{\bibinfo{person}{Trung~Quoc Luong}, \bibinfo{person}{Xinbo Zhang}, \bibinfo{person}{Zhanming Jie}, \bibinfo{person}{Peng Sun}, \bibinfo{person}{Xiaoran Jin}, {and} \bibinfo{person}{Hang Li}.} \bibinfo{year}{2024}\natexlab{}.
\newblock \showarticletitle{Reft: Reasoning with reinforced fine-tuning}.
\newblock \bibinfo{journal}{\emph{arXiv preprint arXiv:2401.08967}} (\bibinfo{year}{2024}).
\newblock


\bibitem[Madaan et~al\mbox{.}(2024)]%
        {madaan2024self}
\bibfield{author}{\bibinfo{person}{Aman Madaan}, \bibinfo{person}{Niket Tandon}, \bibinfo{person}{Prakhar Gupta}, \bibinfo{person}{Skyler Hallinan}, \bibinfo{person}{Luyu Gao}, \bibinfo{person}{Sarah Wiegreffe}, \bibinfo{person}{Uri Alon}, \bibinfo{person}{Nouha Dziri}, \bibinfo{person}{Shrimai Prabhumoye}, \bibinfo{person}{Yiming Yang}, {et~al\mbox{.}}} \bibinfo{year}{2024}\natexlab{}.
\newblock \showarticletitle{Self-refine: Iterative refinement with self-feedback}.
\newblock \bibinfo{journal}{\emph{Advances in Neural Information Processing Systems}}  \bibinfo{volume}{36} (\bibinfo{year}{2024}).
\newblock


\bibitem[o1~Team(2024)]%
        {skyworkopeno12024}
\bibfield{author}{\bibinfo{person}{Skywork o1 Team}.} \bibinfo{year}{2024}\natexlab{}.
\newblock \bibinfo{title}{Skywork-o1 Open Series}.
\newblock \bibinfo{howpublished}{\url{https://huggingface.co/Skywork}}.
\newblock
\urldef\tempurl%
\url{https://huggingface.co/Skywork}
\showURL{%
\tempurl}


\bibitem[Pang et~al\mbox{.}(2024)]%
        {pang2024iterative}
\bibfield{author}{\bibinfo{person}{Richard~Yuanzhe Pang}, \bibinfo{person}{Weizhe Yuan}, \bibinfo{person}{Kyunghyun Cho}, \bibinfo{person}{He He}, \bibinfo{person}{Sainbayar Sukhbaatar}, {and} \bibinfo{person}{Jason Weston}.} \bibinfo{year}{2024}\natexlab{}.
\newblock \showarticletitle{Iterative reasoning preference optimization}.
\newblock \bibinfo{journal}{\emph{arXiv preprint arXiv:2404.19733}} (\bibinfo{year}{2024}).
\newblock


\bibitem[Press et~al\mbox{.}(2022)]%
        {press2022measuring}
\bibfield{author}{\bibinfo{person}{Ofir Press}, \bibinfo{person}{Muru Zhang}, \bibinfo{person}{Sewon Min}, \bibinfo{person}{Ludwig Schmidt}, \bibinfo{person}{Noah~A Smith}, {and} \bibinfo{person}{Mike Lewis}.} \bibinfo{year}{2022}\natexlab{}.
\newblock \showarticletitle{Measuring and narrowing the compositionality gap in language models}.
\newblock \bibinfo{journal}{\emph{arXiv preprint arXiv:2210.03350}} (\bibinfo{year}{2022}).
\newblock


\bibitem[Press et~al\mbox{.}(2023)]%
        {press2023measuring}
\bibfield{author}{\bibinfo{person}{Ofir Press}, \bibinfo{person}{Muru Zhang}, \bibinfo{person}{Sewon Min}, \bibinfo{person}{Ludwig Schmidt}, \bibinfo{person}{Noah~A Smith}, {and} \bibinfo{person}{Mike Lewis}.} \bibinfo{year}{2023}\natexlab{}.
\newblock \showarticletitle{Measuring and Narrowing the Compositionality Gap in Language Models}. In \bibinfo{booktitle}{\emph{Findings of the Association for Computational Linguistics: EMNLP 2023}}. \bibinfo{pages}{5687--5711}.
\newblock


\bibitem[Shao et~al\mbox{.}(2023)]%
        {shao2023enhancing}
\bibfield{author}{\bibinfo{person}{Zhihong Shao}, \bibinfo{person}{Yeyun Gong}, \bibinfo{person}{Yelong Shen}, \bibinfo{person}{Minlie Huang}, \bibinfo{person}{Nan Duan}, {and} \bibinfo{person}{Weizhu Chen}.} \bibinfo{year}{2023}\natexlab{}.
\newblock \showarticletitle{Enhancing retrieval-augmented large language models with iterative retrieval-generation synergy}.
\newblock \bibinfo{journal}{\emph{arXiv preprint arXiv:2305.15294}} (\bibinfo{year}{2023}).
\newblock


\bibitem[Shi et~al\mbox{.}(2023)]%
        {shi2023replug}
\bibfield{author}{\bibinfo{person}{Weijia Shi}, \bibinfo{person}{Sewon Min}, \bibinfo{person}{Michihiro Yasunaga}, \bibinfo{person}{Minjoon Seo}, \bibinfo{person}{Rich James}, \bibinfo{person}{Mike Lewis}, \bibinfo{person}{Luke Zettlemoyer}, {and} \bibinfo{person}{Wen-tau Yih}.} \bibinfo{year}{2023}\natexlab{}.
\newblock \showarticletitle{Replug: Retrieval-augmented black-box language models}.
\newblock \bibinfo{journal}{\emph{arXiv preprint arXiv:2301.12652}} (\bibinfo{year}{2023}).
\newblock


\bibitem[Singh et~al\mbox{.}(2023)]%
        {singh2023beyond}
\bibfield{author}{\bibinfo{person}{Avi Singh}, \bibinfo{person}{John~D Co-Reyes}, \bibinfo{person}{Rishabh Agarwal}, \bibinfo{person}{Ankesh Anand}, \bibinfo{person}{Piyush Patil}, \bibinfo{person}{Xavier Garcia}, \bibinfo{person}{Peter~J Liu}, \bibinfo{person}{James Harrison}, \bibinfo{person}{Jaehoon Lee}, \bibinfo{person}{Kelvin Xu}, {et~al\mbox{.}}} \bibinfo{year}{2023}\natexlab{}.
\newblock \showarticletitle{Beyond human data: Scaling self-training for problem-solving with language models}.
\newblock \bibinfo{journal}{\emph{arXiv preprint arXiv:2312.06585}} (\bibinfo{year}{2023}).
\newblock


\bibitem[Snell et~al\mbox{.}(2024)]%
        {snell2024scaling}
\bibfield{author}{\bibinfo{person}{Charlie Snell}, \bibinfo{person}{Jaehoon Lee}, \bibinfo{person}{Kelvin Xu}, {and} \bibinfo{person}{Aviral Kumar}.} \bibinfo{year}{2024}\natexlab{}.
\newblock \showarticletitle{Scaling llm test-time compute optimally can be more effective than scaling model parameters}.
\newblock \bibinfo{journal}{\emph{arXiv preprint arXiv:2408.03314}} (\bibinfo{year}{2024}).
\newblock


\bibitem[Sutton(2019)]%
        {sutton2019bitter}
\bibfield{author}{\bibinfo{person}{Richard Sutton}.} \bibinfo{year}{2019}\natexlab{}.
\newblock \bibinfo{title}{The Bitter Lesson}.
\newblock
\urldef\tempurl%
\url{http://incompleteideas.net/IncIdeas/BitterLesson.html}
\showURL{%
\tempurl}
\newblock
\shownote{Incomplete Ideas (blog), 13(1):38}.


\bibitem[Sutton and Barto(2018)]%
        {sutton2018reinforcement}
\bibfield{author}{\bibinfo{person}{Richard~S Sutton} {and} \bibinfo{person}{Andrew~G Barto}.} \bibinfo{year}{2018}\natexlab{}.
\newblock \bibinfo{booktitle}{\emph{Reinforcement learning: An introduction}}.
\newblock \bibinfo{publisher}{MIT press}.
\newblock


\bibitem[Tesauro(1995)]%
        {td}
\bibfield{author}{\bibinfo{person}{Gerald Tesauro}.} \bibinfo{year}{1995}\natexlab{}.
\newblock \showarticletitle{Temporal difference learning and TD-Gammon}.
\newblock \bibinfo{journal}{\emph{Commun. ACM}} \bibinfo{volume}{38}, \bibinfo{number}{3} (\bibinfo{date}{March} \bibinfo{year}{1995}), \bibinfo{pages}{58–68}.
\newblock
\showISSN{0001-0782}
\urldef\tempurl%
\url{https://doi.org/10.1145/203330.203343}
\showDOI{\tempurl}


\bibitem[Trivedi et~al\mbox{.}(2022)]%
        {trivedi2022musique}
\bibfield{author}{\bibinfo{person}{Harsh Trivedi}, \bibinfo{person}{Niranjan Balasubramanian}, \bibinfo{person}{Tushar Khot}, {and} \bibinfo{person}{Ashish Sabharwal}.} \bibinfo{year}{2022}\natexlab{}.
\newblock \showarticletitle{MuSiQue: Multihop Questions via Single-hop Question Composition}.
\newblock \bibinfo{journal}{\emph{Transactions of the Association for Computational Linguistics}}  \bibinfo{volume}{10} (\bibinfo{year}{2022}), \bibinfo{pages}{539--554}.
\newblock


\bibitem[Trivedi et~al\mbox{.}(2023)]%
        {trivedi2023interleaving}
\bibfield{author}{\bibinfo{person}{Harsh Trivedi}, \bibinfo{person}{Niranjan Balasubramanian}, \bibinfo{person}{Tushar Khot}, {and} \bibinfo{person}{Ashish Sabharwal}.} \bibinfo{year}{2023}\natexlab{}.
\newblock \showarticletitle{Interleaving Retrieval with Chain-of-Thought Reasoning for Knowledge-Intensive Multi-Step Questions}. In \bibinfo{booktitle}{\emph{Proceedings of the 61st Annual Meeting of the Association for Computational Linguistics (Volume 1: Long Papers)}}. \bibinfo{pages}{10014--10037}.
\newblock


\bibitem[Trung et~al\mbox{.}(2024)]%
        {trung-etal-2024-reft}
\bibfield{author}{\bibinfo{person}{Luong Trung}, \bibinfo{person}{Xinbo Zhang}, \bibinfo{person}{Zhanming Jie}, \bibinfo{person}{Peng Sun}, \bibinfo{person}{Xiaoran Jin}, {and} \bibinfo{person}{Hang Li}.} \bibinfo{year}{2024}\natexlab{}.
\newblock \showarticletitle{{R}e{FT}: Reasoning with Reinforced Fine-Tuning}. In \bibinfo{booktitle}{\emph{Proceedings of the 62nd Annual Meeting of the Association for Computational Linguistics (Volume 1: Long Papers)}}, \bibfield{editor}{\bibinfo{person}{Lun-Wei Ku}, \bibinfo{person}{Andre Martins}, {and} \bibinfo{person}{Vivek Srikumar}} (Eds.). \bibinfo{publisher}{Association for Computational Linguistics}, \bibinfo{address}{Bangkok, Thailand}, \bibinfo{pages}{7601--7614}.
\newblock
\urldef\tempurl%
\url{https://doi.org/10.18653/v1/2024.acl-long.410}
\showDOI{\tempurl}


\bibitem[Wang et~al\mbox{.}(2022)]%
        {wang2022text}
\bibfield{author}{\bibinfo{person}{Liang Wang}, \bibinfo{person}{Nan Yang}, \bibinfo{person}{Xiaolong Huang}, \bibinfo{person}{Binxing Jiao}, \bibinfo{person}{Linjun Yang}, \bibinfo{person}{Daxin Jiang}, \bibinfo{person}{Rangan Majumder}, {and} \bibinfo{person}{Furu Wei}.} \bibinfo{year}{2022}\natexlab{}.
\newblock \showarticletitle{Text embeddings by weakly-supervised contrastive pre-training}.
\newblock \bibinfo{journal}{\emph{arXiv preprint arXiv:2212.03533}} (\bibinfo{year}{2022}).
\newblock


\bibitem[Wang et~al\mbox{.}(2024a)]%
        {wang2024math}
\bibfield{author}{\bibinfo{person}{Peiyi Wang}, \bibinfo{person}{Lei Li}, \bibinfo{person}{Zhihong Shao}, \bibinfo{person}{Runxin Xu}, \bibinfo{person}{Damai Dai}, \bibinfo{person}{Yifei Li}, \bibinfo{person}{Deli Chen}, \bibinfo{person}{Yu Wu}, {and} \bibinfo{person}{Zhifang Sui}.} \bibinfo{year}{2024}\natexlab{a}.
\newblock \showarticletitle{Math-shepherd: Verify and reinforce llms step-by-step without human annotations}. In \bibinfo{booktitle}{\emph{Proceedings of the 62nd Annual Meeting of the Association for Computational Linguistics (Volume 1: Long Papers)}}. \bibinfo{pages}{9426--9439}.
\newblock


\bibitem[Wang et~al\mbox{.}(2023)]%
        {wang2023self}
\bibfield{author}{\bibinfo{person}{Yile Wang}, \bibinfo{person}{Peng Li}, \bibinfo{person}{Maosong Sun}, {and} \bibinfo{person}{Yang Liu}.} \bibinfo{year}{2023}\natexlab{}.
\newblock \showarticletitle{Self-knowledge guided retrieval augmentation for large language models}.
\newblock \bibinfo{journal}{\emph{arXiv preprint arXiv:2310.05002}} (\bibinfo{year}{2023}).
\newblock


\bibitem[Wang et~al\mbox{.}(2024b)]%
        {wang2024multi}
\bibfield{author}{\bibinfo{person}{Zihan Wang}, \bibinfo{person}{Yunxuan Li}, \bibinfo{person}{Yuexin Wu}, \bibinfo{person}{Liangchen Luo}, \bibinfo{person}{Le Hou}, \bibinfo{person}{Hongkun Yu}, {and} \bibinfo{person}{Jingbo Shang}.} \bibinfo{year}{2024}\natexlab{b}.
\newblock \showarticletitle{Multi-step problem solving through a verifier: An empirical analysis on model-induced process supervision}.
\newblock \bibinfo{journal}{\emph{arXiv preprint arXiv:2402.02658}} (\bibinfo{year}{2024}).
\newblock


\bibitem[Wu et~al\mbox{.}(2024)]%
        {wu2024enhancing}
\bibfield{author}{\bibinfo{person}{Zhenyu Wu}, \bibinfo{person}{Qingkai Zeng}, \bibinfo{person}{Zhihan Zhang}, \bibinfo{person}{Zhaoxuan Tan}, \bibinfo{person}{Chao Shen}, {and} \bibinfo{person}{Meng Jiang}.} \bibinfo{year}{2024}\natexlab{}.
\newblock \showarticletitle{Enhancing Mathematical Reasoning in LLMs by Stepwise Correction}.
\newblock \bibinfo{journal}{\emph{arXiv preprint arXiv:2410.12934}} (\bibinfo{year}{2024}).
\newblock


\bibitem[Xie et~al\mbox{.}(2024)]%
        {xie2024monte}
\bibfield{author}{\bibinfo{person}{Yuxi Xie}, \bibinfo{person}{Anirudh Goyal}, \bibinfo{person}{Wenyue Zheng}, \bibinfo{person}{Min-Yen Kan}, \bibinfo{person}{Timothy~P Lillicrap}, \bibinfo{person}{Kenji Kawaguchi}, {and} \bibinfo{person}{Michael Shieh}.} \bibinfo{year}{2024}\natexlab{}.
\newblock \showarticletitle{Monte Carlo Tree Search Boosts Reasoning via Iterative Preference Learning}.
\newblock \bibinfo{journal}{\emph{arXiv preprint arXiv:2405.00451}} (\bibinfo{year}{2024}).
\newblock


\bibitem[Xu et~al\mbox{.}(2024)]%
        {xu2024recomp}
\bibfield{author}{\bibinfo{person}{Fangyuan Xu}, \bibinfo{person}{Weijia Shi}, {and} \bibinfo{person}{Eunsol Choi}.} \bibinfo{year}{2024}\natexlab{}.
\newblock \showarticletitle{{RECOMP}: Improving Retrieval-Augmented {LM}s with Context Compression and Selective Augmentation}. In \bibinfo{booktitle}{\emph{The Twelfth International Conference on Learning Representations}}.
\newblock
\urldef\tempurl%
\url{https://openreview.net/forum?id=mlJLVigNHp}
\showURL{%
\tempurl}


\bibitem[Yang et~al\mbox{.}(2018)]%
        {yang2018hotpotqa}
\bibfield{author}{\bibinfo{person}{Zhilin Yang}, \bibinfo{person}{Peng Qi}, \bibinfo{person}{Saizheng Zhang}, \bibinfo{person}{Yoshua Bengio}, \bibinfo{person}{William Cohen}, \bibinfo{person}{Ruslan Salakhutdinov}, {and} \bibinfo{person}{Christopher~D Manning}.} \bibinfo{year}{2018}\natexlab{}.
\newblock \showarticletitle{HotpotQA: A Dataset for Diverse, Explainable Multi-hop Question Answering}. In \bibinfo{booktitle}{\emph{Proceedings of the 2018 Conference on Empirical Methods in Natural Language Processing}}. \bibinfo{pages}{2369--2380}.
\newblock


\bibitem[Yao et~al\mbox{.}(2023)]%
        {yao2023react}
\bibfield{author}{\bibinfo{person}{Shunyu Yao}, \bibinfo{person}{Jeffrey Zhao}, \bibinfo{person}{Dian Yu}, \bibinfo{person}{Nan Du}, \bibinfo{person}{Izhak Shafran}, \bibinfo{person}{Karthik Narasimhan}, {and} \bibinfo{person}{Yuan Cao}.} \bibinfo{year}{2023}\natexlab{}.
\newblock \showarticletitle{ReAct: Synergizing Reasoning and Acting in Language Models}. In \bibinfo{booktitle}{\emph{International Conference on Learning Representations (ICLR)}}.
\newblock


\bibitem[Yue et~al\mbox{.}(2024)]%
        {yue2024inference}
\bibfield{author}{\bibinfo{person}{Zhenrui Yue}, \bibinfo{person}{Honglei Zhuang}, \bibinfo{person}{Aijun Bai}, \bibinfo{person}{Kai Hui}, \bibinfo{person}{Rolf Jagerman}, \bibinfo{person}{Hansi Zeng}, \bibinfo{person}{Zhen Qin}, \bibinfo{person}{Dong Wang}, \bibinfo{person}{Xuanhui Wang}, {and} \bibinfo{person}{Michael Bendersky}.} \bibinfo{year}{2024}\natexlab{}.
\newblock \showarticletitle{Inference scaling for long-context retrieval augmented generation}.
\newblock \bibinfo{journal}{\emph{arXiv preprint arXiv:2410.04343}} (\bibinfo{year}{2024}).
\newblock


\bibitem[Zhang et~al\mbox{.}(2024)]%
        {zhang2024openrft}
\bibfield{author}{\bibinfo{person}{Yuxiang Zhang}, \bibinfo{person}{Yuqi Yang}, \bibinfo{person}{Jiangming Shu}, \bibinfo{person}{Yuhang Wang}, \bibinfo{person}{Jinlin Xiao}, {and} \bibinfo{person}{Jitao Sang}.} \bibinfo{year}{2024}\natexlab{}.
\newblock \showarticletitle{OpenRFT: Adapting Reasoning Foundation Model for Domain-specific Tasks with Reinforcement Fine-Tuning}.
\newblock \bibinfo{journal}{\emph{arXiv preprint arXiv:2412.16849}} (\bibinfo{year}{2024}).
\newblock


\bibitem[Zhao et~al\mbox{.}(2024)]%
        {zhao2024marco}
\bibfield{author}{\bibinfo{person}{Yu Zhao}, \bibinfo{person}{Huifeng Yin}, \bibinfo{person}{Bo Zeng}, \bibinfo{person}{Hao Wang}, \bibinfo{person}{Tianqi Shi}, \bibinfo{person}{Chenyang Lyu}, \bibinfo{person}{Longyue Wang}, \bibinfo{person}{Weihua Luo}, {and} \bibinfo{person}{Kaifu Zhang}.} \bibinfo{year}{2024}\natexlab{}.
\newblock \showarticletitle{Marco-o1: Towards open reasoning models for open-ended solutions}.
\newblock \bibinfo{journal}{\emph{arXiv preprint arXiv:2411.14405}} (\bibinfo{year}{2024}).
\newblock


\end{thebibliography}

\end{document}